\pdfoutput=1

\documentclass[11pt]{article}

\usepackage[final]{acl}

\usepackage{times}
\usepackage{latexsym}
\usepackage[T1]{fontenc}
\usepackage[utf8]{inputenc}
\usepackage[nameinlink]{cleveref}
\usepackage{changepage}
\usepackage{booktabs}
\usepackage{microtype}
\usepackage{tabularx}
\usepackage{inconsolata}
\usepackage{array, makecell, booktabs}
\usepackage{multirow}
\usepackage{graphicx}
\usepackage{paralist}

\title{Limitations of Religious Data and the Importance of the Target Domain: Towards Machine Translation for Guinea-Bissau Creole}

\author{
 \textbf{Jacqueline Rowe\textsuperscript{1}},
 \textbf{Edward Gow-Smith\textsuperscript{2}},
 \textbf{Mark Hepple\textsuperscript{2}}
\\
\\
 \textsuperscript{1}University of Edinburgh \\
 \textsuperscript{2}University of Sheffield
\\
 \small{
  \href{mailto:email@domain}{jacqueline.rowe@ed.ac.uk}
 }
}

\begin{document}
\maketitle
\begin{abstract}
We introduce a new dataset for machine translation of Guinea-Bissau Creole (Kiriol), comprising around 40 thousand parallel sentences to English and Portuguese. This dataset is made up of predominantly religious data (from the Bible and texts from the Jehovah's Witnesses), but also a small amount of general domain data (from a dictionary). This mirrors the typical resource availability of many low resource languages. We train a number of transformer-based models to investigate how to improve domain transfer from religious data to a more general domain. We find that adding even 300 sentences from the target domain when training substantially improves the translation performance, highlighting the importance and need for data collection for low-resource languages, even on a small-scale. We additionally find that Portuguese-to-Kiriol translation models perform better on average than other source and target language pairs, and investigate how this relates to the morphological complexity of the languages involved and the degree of lexical overlap between creoles and lexifiers. Overall, we hope our work will stimulate research into Kiriol and into how machine translation might better support creole languages in general.
\end{abstract}

\section{Introduction}
The proliferation of datasets and models for a wide range of the world's languages has significantly expanded the coverage of machine translation (MT) technologies in recent years. Nevertheless, despite their often significant numbers of speakers, creole languages are digitally underrepresented and have minimal representation in MT technologies \cite{robinson-etal-2024-kreyol,lent2024creoleval}. This is, at least in part, because creoles have historically been subject to stigmatisation, particularly in administrative and educational settings \cite{siegel1999creoles, wigglesworth2013creole}. For many creoles, like other low-resource languages, translations of religious texts form the largest parallel digital resources available \cite{siddhant2022towards, lent-etal-2021-language}, but language technologies trained on religious data alone have been shown to have limited applicability in non-religious contexts \cite{haddow2022survey}. Religious texts not only focus on theological concepts but are also rooted in specific socio-cultural contexts which may not align with those of the region where an MT model would be used \cite{kho2024some}. Using religious texts for NLP applications and models may also pose particular harms when implemented in practice \cite{hutchinson2024modeling}, making it especially important to evaluate models trained on religious data with a view to the downstream domain.


In this work, we look at Guinea-Bissau Creole (also known as Kiriol\footnote{\url{https://www.ethnologue.com/language/pov/}}), a Portuguese creole spoken primarily in Guinea-Bissau. With approximately 350,000 native speakers and 1.5 million L2 speakers, it is used by over 90\% of the population of Guinea-Bissau and is the de facto language of national identity \citep[p.192]{kohl2016limitations}. Despite this, digital support for Kiriol is low, with religious texts and dictionary resources being virtually the only public digital material in Kiriol currently available to our knowledge. Thus, our aims are twofold: to introduce a dataset for Kiriol which provides resources for building MT models, and to investigate how this data can be best utilised to train translation models for general (non-religious) domains.




Our contributions are as follows: 
\begin{compactitem}
    \item A new dataset of nearly 40,000 parallel sentences in Kiriol, English and Portuguese;\footnote{To prevent irresponsible use of this low-resource data for inclusion in large multilingual models without consent, and also to adhere to the copyright terms of the publishers of the dataset sources, we make our dataset available to academic researchers only upon request. Please contact the lead author for license agreement and access.} 
    \item A series of from-scratch models trained for Kiriol-English, English-Kiriol, Kiriol-Portuguese and Portuguese-Kiriol translation, serving as baselines for future work;
    \item Demonstration of how to improve cross-domain performance by adding small amounts of cross-domain data to religious training sets;
    \item Insights into how the morphological simplicity of creole languages may disadvantage their representations when sharing tokenisers with more morphologically complex languages;
    \item Evidence that lexical overlap between creoles and lexifier languages improves MT performance where combined tokenisers and shared embedding layers are properly utilised.  
\end{compactitem}

\section{Related Work}
\subsection{Creole MT}
The growing body of research on creole MT includes languages such as Haitian and Jamaican Kreyol \citep{robinson-etal-2022-data}, Mauritian Creole \citep{dabre-etal-2014-anou}, Nigeran Pidgin \citep{ahia2020towards}, Singlish \citep{lent-etal-2021-language} and Sranan Tongo \citep{zwennicker2022towards}. While a wider range of creoles are receiving more visibility and attention within NLP more generally, to our knowledge the only existing efforts towards MT for Guinea-Bissau creole specifically are that of \citet{robinson-etal-2024-kreyol}, who include limited Kiriol data in their multilingual creole corpus and set of from-scratch and finetuned multilingual creole models. Their dataset of 480 parallel English-Kiriol sentences is taken from an unnamed educational resource; this provides a good foundation, but is limited in size and concentrated on a single domain, and the dataset's non-standard Kiriol orthography means that evaluation on this set may not be indicative of translation performance on standard Kiriol. In this work, we focus on MT between Kiriol and both Portuguese and English; English because of its global coverage, and Portuguese because it is the official language of Guinea-Bissau, despite the fact that many Bissau-Guineans do not speak it fluently \citep[p.167]{kohl2018creole}.

\subsection{Leveraging Bible data for MT}
Given that parallel texts for many languages are limited to narrow-domain religious texts \citep{siddhant2022towards}, others have investigated how best to leverage or augment such data for low-resource MT. For example, \citet{liu-etal-2021-usefulness} investigate how dictionary data could be used to improve models trained with Bible data for Basque-English and Navajo-English, finding that adding domain-general data to the Bible did not improve performance on their Bible test set. Their use of the Bible (rather than a domain-general set) for evaluation limits the applicability of their findings to other translation domains. The same is true for \citet{mueller-etal-2020-analysis}, who train multilingual translation models covering over 1,000 languages on Bible data and test them on held out Bible data. \citet{marashian-etal-2025-priest} explore how adding dictionary data to the Bible can assist domain adaptation of MT models from religious to non-religious domains, working with English and five simulated low-resource languages. Yet, they also include up to 200k monolingual source-side sentences in the target domains, making this a much higher-resource context than our work.

\section{Dataset}
\label{section:Dataset}
Here we introduce our dataset, taken from three online resources: Bible.com,\footnote{Available at \url{https://www.bible.com}, accessed 13 February 2025.} monthly publications from the Jehovah's Witnesses (JW) website,\footnote{Available at \url{https://www.jw.org/en/library/magazines/} accessed 13 February 2025. Copyright © 2025 Watch Tower Bible and Tract Society of Pennsylvania.}, and a Portuguese-Kiriol bilingual dictionary.\footnote{Available at \url{https://www.editora.ufpb.br/sistema/press5/index.php/UFPB/catalog/download/705/941/8096-1?inline=1}, accessed 13 February 2025. Copyright © 2021 – UFPB Publishing.} We extract parallel aligned Bible verses from the Almeida Revista e Corrigida (Portuguese),\footnote{Copyright © 2001 Sociedade Bíblica de Portugal} the New International Version (English) \footnote{Copyright © 1973, 1978, 1984, 2011 by Biblica, Inc.} and the Traduson Antigu (Kiriol).\footnote{Copyright © 1993-2020, Instituto de Tradução e Alfabetização.} We extract English, Portuguese and Kiriol paragraphs from online editions of JW's Watchtower magazine (WT), a monthly Bible study resource, and a small number of paragraphs from the JW monthly article series ``How Your Donations Are Used''\footnote{Available at \url{https://www.jw.org/en/library/series/how-your-donations-are-used/}, accessed 13 February 2025.}. Both of the JW resources are available in a wide range of low-resource languages, and while they are religious-themed, they use more informal language than the biblical texts and include discussion of more general topics such as finance, communication, transport, relationships, and technology. As such, we describe these as semi-religious data. For both JW resources, we align the extracted paragraphs using HTML tags, meaning these entries can consist of multiple sentences. 

Finally, we extract 6,902 vocabulary items in Kiriol and Portuguese from the bilingual dictionary. We use NLLB \citep{nllb2024scaling} to automatically translate the Portuguese vocabulary items to English, removing any translations that are more than two words long to remove likely errors. This creates a three-way parallel lexicon of 1,983 lexical items. We also extract 1,603 Kiriol and Portuguese glosses from the dictionary, using NLLB to translate the Portuguese sentences into English, then manually checking and correcting the English translations with reference to the original Kiriol glosses. The dictionary sentences are not from the religious domain, but cover a wide range of day-to-day topics such as food, family, work and lifestyle. We describe this as `general domain' data. 

In total, we have 38,578 parallel sentences (see \Cref{tab:datasetsize}) and 1,983 vocabulary items in all three languages. There is a small overlap between the dictionary sentences and dictionary items, with 403 of the vocabulary items appearing in the set of 1,603 gloss sentences. While the vocabularies and topics differ between religious and general domain sources, the orthography and punctuation style is consistent across datasets. The dictionary gloss sentences are designed to be simple to aid language learning, and as such they are much shorter on average than the sentences taken from the other resources. Sample sentences are shown in \Cref{tab:sample_sentences}.


\begin{table}[h]
\small
\centering
\begin{tabular}{llr}
\toprule
Source & Domain & \# Sentences \\
\midrule
Bible & Religious & 29,876\\
\hspace{3mm} (Old Testament) & & (22,220) \\
\hspace{3mm} (New Testament) & & (7,656) \\
JW WT series & Semi-Religious & 6,880 \\
JW Donations series & Semi-Religious & 219 \\
Bilingual dictionary & General & 1,603 \\
\midrule
All & & 38,578 \\
\bottomrule
\end{tabular}\\
\caption{Number of sentences collected from each data source. This does not include the 1,983 lexical items also collected from the dictionary.}
\label{tab:datasetsize}
\end{table}

\begin{table*}[h]
\small
\centering
\begin{tabular}{p{2cm}p{4cm}p{4cm}p{4cm}}
\toprule
Source & Kiriol & Portuguese & English \\
\midrule
Bible & Bo jubi pa kacus di seu; e ka ta sumia, e ka ta kebra, e ka ta junta na bemba, ma bo Pape ku sta na seu i ta alimenta elis. Nta abos bo ka mas bali di ki kacus? & Considerai os corvos, que nem semeiam, nem segam, nem têm despensa nem celeiro, e Deus os alimenta; quanto mais valeis vós do que as aves? & Look at the birds of the air; they do not sow or reap or store away in barns, and yet your heavenly Father feeds them. Are you not much more valuable than they? \\
\vspace{0.3cm}\\
JW WT series & Bardadi di Biblia i suma un speliu. Ora ku no lei i studa Biblia, no pudi oja kal tipu di algin ku no sedu pa dentru, i na kal aria ku no pirsisa di minjoria. & A verdade da Bíblia é como um espelho. Quando lemos e estudamos a Bíblia, conseguimos ver o tipo de pessoa que realmente somos e em que áreas precisamos de melhorar. & Bible truth is like a mirror. When we look into it, we can see what we really are on the inside and where we need to make improvements. \\
\vspace{0.3cm}\\
\makecell[tl]{JW Donations \\series} & ma gosi, manga di no irmons ku irmas pudi baŝa publikason diẑital mesmu sin internet! kuma kes i pusivel? & no entanto, agora, muitos irmãos conseguem descarregar publicações digitais mesmo sem ligação à internet! como é que isso é possível? & nevertheless, many of our brothers and sisters can now download digital publications even without an internet connection! how is this possible? \\
\vspace{0.3cm}\\

\makecell[tl]{Bilingual\\dictionary} & Ña dona sufri ataki di korson. & Meu avô sofreu um ataque do coração. & My grandfather suffered a heart attack. \\
\vspace{0.3cm}\\
\makecell[tl]{Bilingual\\lexicon} & armasen & armazém & warehouse \\
\bottomrule
\end{tabular}
\caption{Sample entries from each dataset}
\label{tab:sample_sentences}
\end{table*}

We construct our training and test datasets to best reflect real-world translation scenarios. The bilingual dictionary sentences cover a broader range of topics in the general domain. We therefore take 1,000 of these sentences to comprise our out-of-domain test set. We construct a validation set of 500 JW WT sentences and 500 Bible sentences (1,000 total). This leaves us with 36,578 training sentences from the Bible, JW WT, JW Donations and bilingual dictionary, covering a range of domains but with mostly religious data.

\section{Models}
We describe here the training procedure for all our translation models, which are trained from-scratch using an encoder-decoder transformer architecture \cite{vaswani2017attention}. This is implemented using the Eole toolkit,\footnote{\url{https://github.com/eole-nlp/eole}} an updated version of the Open-NMT toolkit \cite{klein2018opennmt}. For each model, we train a byte-pair encoding tokeniser \cite{sennrich2015neural} with a vocabulary size of 10k (unless otherwise specified), implemented in Sentence Piece \cite{kudo2018subword}. In each experiment, the tokeniser is trained on the same data as used for the model training. We then train models for up to 20k steps using the Adam optimiser \cite{kingma2017adammethodstochasticoptimization}, validating with BLEU \cite{papineni2002bleu} every 1,000 steps with an early stopping of 5 validations: the best model is then evaluated on the test set.\footnote{We evaluate our models using the SacreBLEU \cite{post-2018-call} signature: \texttt{nrefs:1|case:mixed|eff:no|tok:13a\\|smooth:exp|version:2.4.3'}.} For each experiment, we run five seeds on all four language directions (Kiriol-Portuguese, Kiriol-English, Portuguese-Kiriol, English-Kiriol). This setup improves the robustness of the results, even where BLEU scores are low. For each experiment, we report the average test and validation BLEU scores over the five seeds in \Cref{app:scores}. A full list of model parameters is found in \Cref{app:params}.

\section{Experiments}
We run a series of experiments to explore the utility of the different datasets for training models from scratch to generalise across domains, and to investigate differences in performance between translation directions and language pairs. 

\subsection{Training with only Religious Data}
\label{section:4.1}
\begin{figure*}[h!]
    \centering
    \includegraphics[width=1\linewidth]{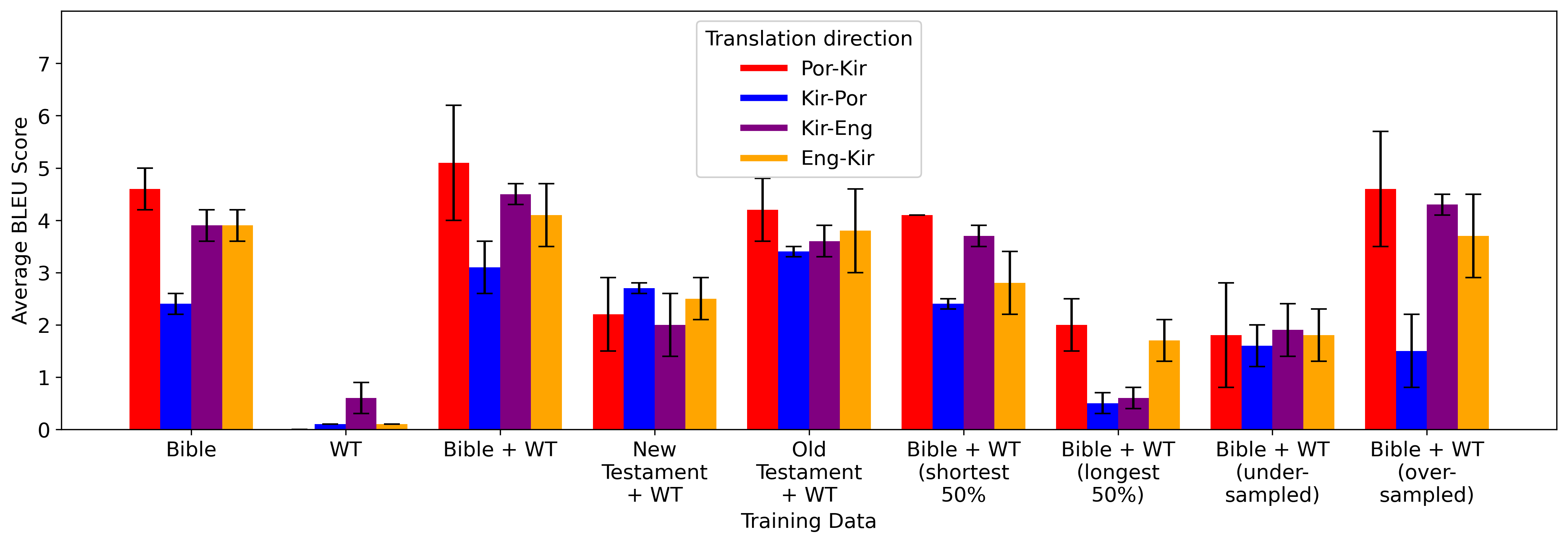}
    \caption{Average performance of Portuguese-Kiriol, Kiriol-Portuguese, Kiriol-English and English-Kiriol models trained on different portions of Bible and WT data when used to translate test set of 1,000 domain-general dictionary sentences. Standard errors across model sets are shown with error bars.}
    \label{fig:set1}
\end{figure*}

We first explore the utility of the Bible and WT data (the largest two resources) for training translation models that can generalise to unseen domains, which previous work has questioned \cite{gow-smith-snchez-villegas-2023-sheffields,vazquez-etal-2021-helsinki}. We train from-scratch translation models on different combinations of Bible and WT data, exploring whether different sampling methods improve the performance of the models on the domain-general dictionary test set. Average BLEU scores and standard errors on the test set are shown in \Cref{fig:set1}; the standard errors are large in most cases due to the small size and limited domain of these training datasets. \Cref{tab:full_results_1} in \Cref{app:scores} shows scores and standard errors on all evaluation sets, along with the training dataset sizes in each condition. 

Our first experiments train models just with the Bible, just with WT, and then with both; we find that the combination models show modest improvements on the Bible-only models, while the models trained with WT only perform extremely poorly on all evaluation sets, likely because of the small size of the dataset overall. Next, as different chapters of the Bible come from a variety of authors, eras, styles and contexts, we explore whether selective sampling of Bible texts might improve performance. Specifically, we combine first the Old Testament (OT) and then the New Testament (NT),\footnote{The OT contains more poetry, genealogies and legal language, covering key moments of Jewish history and philosophy, while the NT consists of parables, letters and gospel accounts of the life of Jesus and his disciples and the concerns of the early Christian church.} with the WT data, and train new models on these mixtures. On average, the OT \& WT models perform better on the test set, likely because the OT is much larger than the NT. We next sample the shortest and longest 50\% of sentences across both the Bible and WT datasets, to explore whether sentence length impacts model performance on the test set, which consists of mostly short sentences. The models trained on the shorter sentences perform better on the test set compared to the models trained on the longer sentences, although we note that the latter perform better on the validation sets (see \Cref{tab:full_results_1}).

Finally, we explore the impact of balancing the training dataset across both Bible and WT sources, first by randomly sampling the same number of sentences from the Bible as are in the WT dataset, and secondly by oversampling sentences from the WT dataset until its size matches that of the whole Bible. The undersampled models perform quite poorly, while the oversampled models perform well but are still outperformed by some of the other training data setups, despite the fact that they are trained on the largest dataset in the group. 

Out of all training setups, the simple combination of Bible and WT data results in models with the highest average performance on the test set, but the high variability in results makes it challenging to draw definitive conclusions. Furthermore, despite a modest amount of data (36k sentences), the results are still poor (average of 4.23 BLEU across all language directions), demonstrating the limitations of models trained solely on religious data when applied to a domain-general test set.


\subsection{Training With Additional General-Domain Data}
\label{section:4.2}

\begin{figure*}[h!]
\centering
\includegraphics[width=1\linewidth]{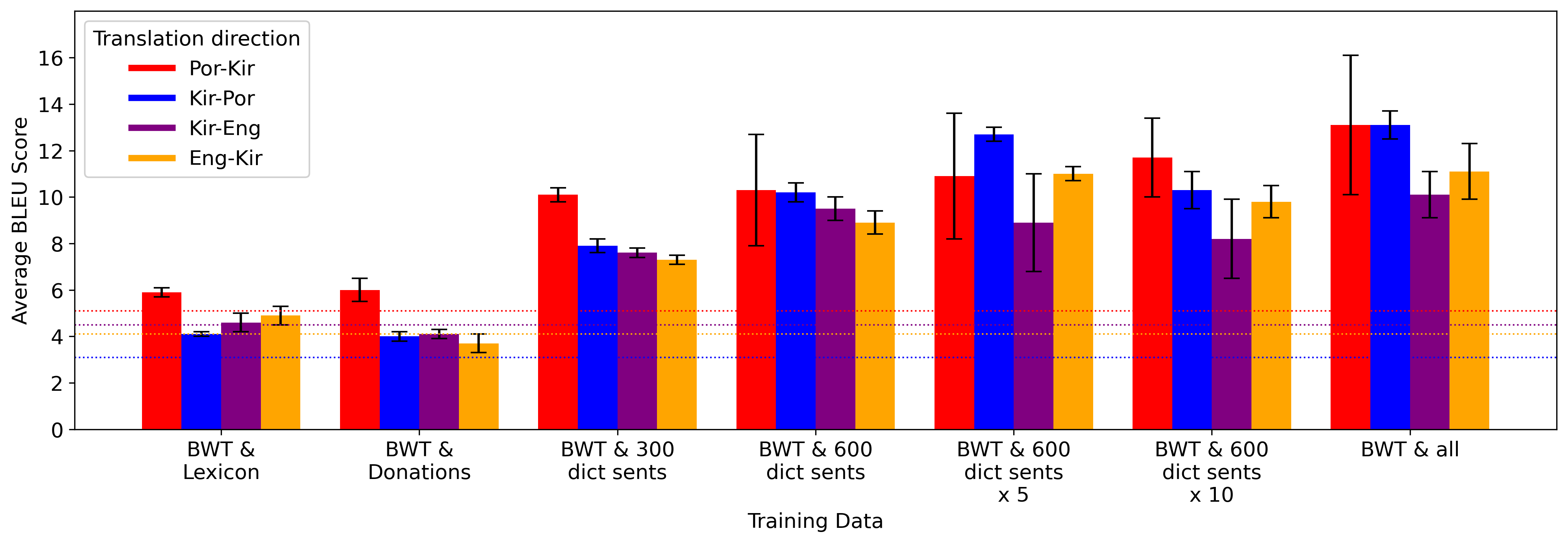}
\caption{Average performance of Portuguese-Kiriol, Kiriol-Portuguese, Kiriol-English and English-Kiriol models trained on Bible, WT and different combinations of domain-general data when used to translate test set of 1,000 domain-general dictionary sentences. Standard errors across model sets are shown with error bars, and the baseline average performance of models trained only on Bible and WT data is shown with dotted lines.}
\label{fig:set2}
\end{figure*}

Given these results, we explore whether adding a small amount of general-domain data to the religious training data can improve model performance. We take the best-performing setup from the previous step (Bible and WT) as a baseline, but add different samples of additional, non-religious data from the Donations series and the bilingual lexicon, resources which are likely available in many other low-resource languages. We present the average performance of each model set on the test set in \Cref{fig:set2}, with dataset sizes and average scores on all evaluation sets in \Cref{tab:full_results_2} in \Cref{app:scores}.


First, we add the 1,983 lexical items from the dictionary (284 of which also appear in the 1,000 dictionary sentences split out for the test set) and then the 219 sentences from the JW Donations series. These models do not show considerable improvements over the baseline average model performance scores, and adding the Donations sentences actually results in small decreases in average performance for models translating between English and Kiriol, although this may be random variation. 

Next, we add small amounts of parallel data from the target domain (the dictionary), based on the fact that it is often possible, in a real-world low-resource translation scenario, to collect a few hundred manually-translated sentences in the domain of interest. Given that we have 603 dictionary gloss sentences which are not included in the test set, we train models by adding 300 dictionary sentences, 600 dictionary sentences, and 600 dictionary sentences sampled five times and ten times. The results show, firstly, that adding these very small amounts of target domain data drastically improves the BLEU scores on the test set, by between 4.0 and 6.7 BLEU. Adding 600 sentences oversampled 5 times produces best results on the test set, indicating that very small amounts of target domain data can be of high utility to model training and subsequent performance on the target domain, even when sampled multiple times. These improvements begin to decline again for most language directions when we oversample 10 times, however, suggesting a trade-off between the proportion of the training dataset that is similar to the target domain, and the degree of repetition in the training data. 

Finally, we combine all of these additional sources and again oversample the 600 dictionary sentences 5 times, resulting in an overall dataset of 40,958 training examples. This dataset produces the best performing models overall, with an average BLEU of 11.9 on the test set across all language directions; but as before, there is still considerable variation in the results. 

\begin{figure}[h!]
    \centering
    \includegraphics[width=1\linewidth]{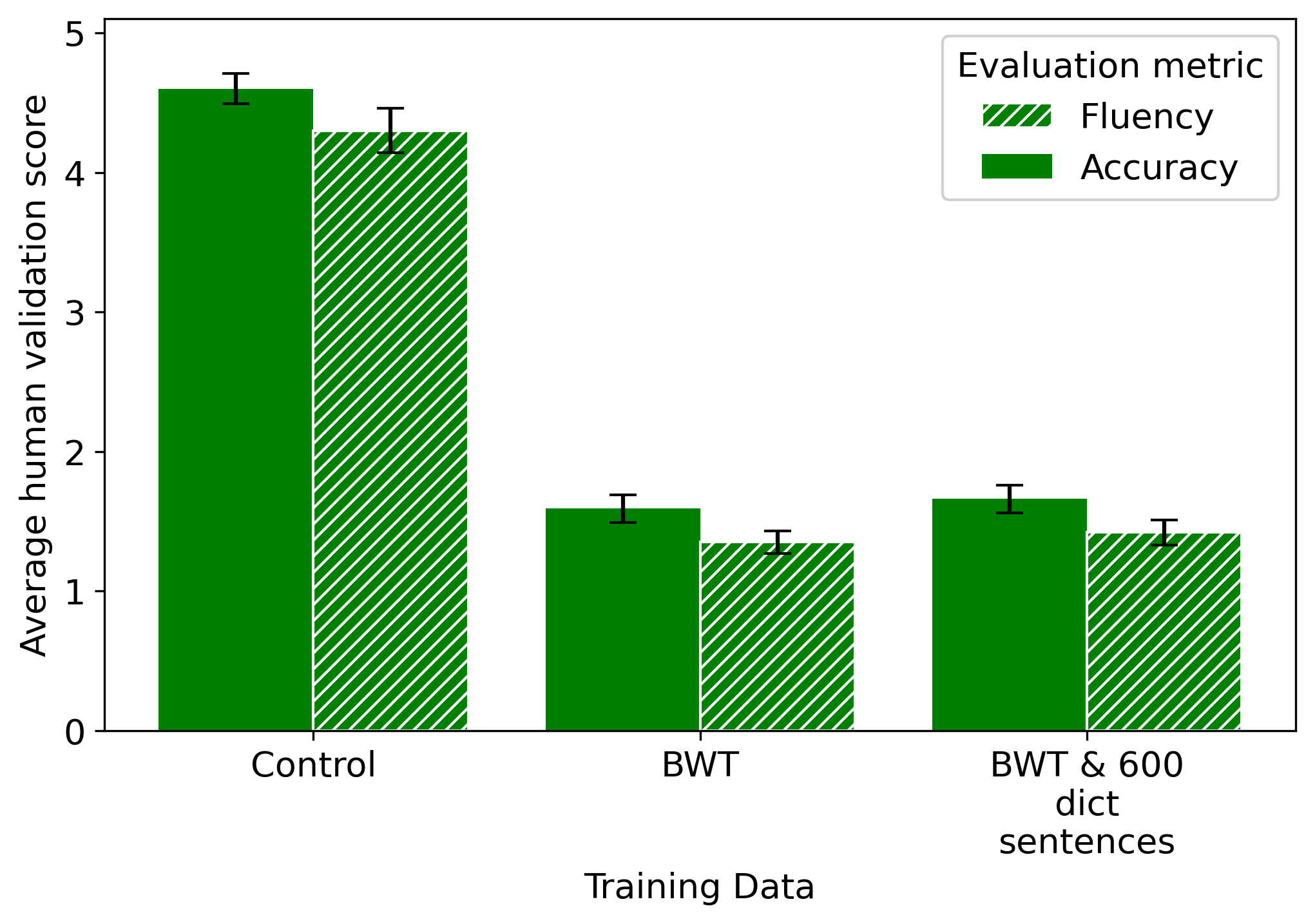}
    \caption{Average scores across all language directions of human judgements for accuracy (solid) and fluency (hatched) of translated sentences from the reference sets (control) and from models trained on Bible and WT data (BWT) and Bible, WT and 600 dictionary sentences. Standard errors across all judgements for each condition are shown with error bars.}
    \label{fig:validation_scores}
\end{figure}

To explore the robustness of the improvements in translation performance on the test set, we conducted a small-scale human validation experiment with a native speaker (NS) of Kiriol who is also fluent in English and Portuguese. The NS was asked to rate sentences translated in all language directions for fluency and accuracy on a scale of 1 to 5, following \citet{koehn-monz-2006-manual}. In each language direction, the NS rated 10 sentences from the reference sets (to serve as a control), and 25 sentences translated by both the models trained on Bible and WT data and the models trained on Bible, WT and 600 dictionary sentences. The average results across all language directions are shown in \Cref{fig:validation_scores}; while there is a very slight increase in average scores for accuracy and fluency for the models trained on data including the 600 dictionary sentences, this is not consistent across all language translation directions and overall the scores across the two conditions are very similar. We suspect this is because the quality of translations produced by models both with and without small amounts of target-domain data is still too low to show meaningful improvements in a human judgement task, even though there are notable differences in BLEU score. A breakdown of the human validation scores by language can be found in \Cref{fig:human_val_2} in \Cref{app:scores}, as well as a summary of the costs and instructions given for the validation task in \Cref{app:hval}.

\subsection{Impact of Translation Direction on Performance}
\label{section:4.3}
In both previous sets of experiments, Portuguese-Kiriol models showed the highest performance across most training data setups  (\Cref{fig:set1}, \Cref{fig:set2}). The performance of MT systems has been shown to depend on the characteristics of the languages being translated, including historical relatedness and vocabulary overlap \citep{birch2008predicting} and  morphological complexity  \citep{koehn2005europarl, park2021morphology, cotterell2018all}. Creoles are typically morphologically simple languages \citep{farquharson2007creole} that share vocabulary with their lexifier languages, making these particularly interesting areas of investigation in an MT context. While the higher performance of Portuguese-Kiriol models might suggest greater language similarity or lexical overlap between the two languages, models trained on the opposite direction from Kiriol to Portuguese do not always show the same advantages, and display higher variability in scores.

\subsubsection{Intrinsic Analysis}
To explore this, we first investigate how the vocabularies of Kiriol, English, and Portuguese might impact tokenisation efficiency. Using all available training data (as described for the best-performing models in \cref{section:4.2}), we train shared and separate tokenisers for each language direction with vocabulary sizes of 10k, 20k, and 30k. For the models with separate tokenisers, half the vocabulary is allocated to each of the source and target languages, keeping the overall number of tokens consistent across shared and separate conditions. 

We first count the number of overlapping tokens across separate tokenisers for each language pair, and calculate the average lengths of these overlapping tokens (\Cref{tab:overlapping_vocab}). We find, firstly, that for the separate Kiriol, Portuguese and English tokenisers, there is a much higher degree of overlap between the Kiriol and Portuguese vocabularies than between the Kiriol and English vocabularies. This effect persists across all vocabulary sizes, and the tokens shared by Portuguese and Kiriol are on average longer than those shared by English and Kiriol. This is strong evidence for a greater degree of lexical overlap between Kiriol and Portuguese than Kiriol and English, consistent with what we know about creole genesis \citep{mufwene1996founder}.

\begin{table}[h]
\centering
\small
\begin{tabular}{lccc}
\toprule
& \multirow{2}{*}{\makecell{Vocab. \\size}} &  \\
& & Kir-Eng & Kir-Por \\
\midrule
\multirow{3}{*}{\makecell{\# overlapping tokens \\ between source and \\ target vocabularies}} & 10k & 689 & 1119 \\
& 20k & 1306  & 1982 \\
& 30k & 2149 & 2890 \\
\midrule
\multirow{3}{*}{\makecell{Average length of \\overlapping tokens}} & 10k & 3.04 & 3.51 \\
& 20k & 3.50 & 3.95  \\
& 30k & 3.99 & 4.25 \\

\bottomrule
\end{tabular}
\caption{Number of overlapping tokens in tokeniser vocabularies when separate tokenisers are trained for English, Kiriol and Portuguese using the best combination of training data from Section 5.2.}
\label{tab:overlapping_vocab}
\end{table}

We also explore the number of overlapping tokens in the combined tokenisers' vocabularies by first applying these tokenisers to each language's training data and calculating the how many unique tokens are present in each tokenised dataset. We then add the numbers of unique tokens from both languages together and subtract the size of the vocabulary (10k, 20k or 30k), to calculate how many tokens must be shared between both datasets given the overall vocabulary size. We also calculate the number of tokens from the combined tokeniser which are unique to each language, by subtracting the number of shared tokens from the number of unique tokens present in each language's tokenised training data. The results (\Cref{tab:tokens}) show again that there is a greater degree of overlap between Kiriol and Portuguese compared to Kiriol and English; for example, with vocabulary size of 10k tokens, 3.5k are shared between Kiriol and Portuguese but only 2.6k shared between Kiriol and English. The results also show that Kiriol has fewer unique tokens than either of the other languages across all vocabulary sizes, indicating that more of the combined vocabulary space is allocated to tokens specific to English or Portuguese than to Kiriol.

\begin{table}[h]
\centering
\small
\begin{tabular}{lcccc}
\toprule
& & \multicolumn{3}{c}{Kir-Eng} \\
& \makecell{Vocab. \\size} & Shared & Kir & Eng \\
\midrule
\multirow{3}{*}{\makecell{\# shared and \\ unique tokens \\ across languages}}& 10k & 2649  & 3025  & 4326    \\
& 20k & 3490  & 6691  & 9819   \\
& 30k & 3408  & 11142 & 15450 \\
\midrule
\midrule
& & \multicolumn{3}{c}{Kir-Por} \\
& \makecell{Vocab. \\size} & Shared & Kir & Por \\
\midrule
\multirow{3}{*}{\makecell{\# shared and \\ unique tokens \\ across languages}}& 10k & 3493  & 2448  & 4059 \\ 
& 20k & 4784  & 5336  & 9880 \\
& 30k & 5195  & 8658  & 16147 \\
\bottomrule
\end{tabular}
\caption{Number of overlapping and separate tokens in all training datasets when tokenised by combined tokenisers trained for Kiriol-English and Kiriol-Portuguese using the best combination of training data from Section 5.2.}
\label{tab:tokens}
\end{table}

To investigate these differences further, we compare the fertility of the separate and combined tokenisers on the training datasets in each language, dividing the number of tokens in tokenised datasets by the number of words in untokenised datasets \cite{rust-etal-2021-good}. We expected combined tokenisers to result in better (lower) fertility scores than their equivalent separate tokenisers, as tokens common to both languages occupy only one space in a combined tokeniser vocabulary but two spaces across two separate ones. We show the differences in fertility scores between shared and separate conditions on all language pairs in \Cref{tab:fertility}. We see that the combined tokenisers for both Kiriol-English and Kiriol-Portuguese perform better than separate English-only and Portuguese-only tokenisers on the English and Portuguese data, and this gain is larger for the Portuguese data where we know there is a higher degree of token overlap between Kiriol and Portuguese data (\Cref{tab:overlapping_vocab}). However, when the tokenisers are applied to the Kiriol data, the separate tokenisers are as good as, or slightly better than, the combined ones. As above, this indicates that the Kiriol data is somehow disadvantaged in the combined tokeniser setup, even despite the potential efficiencies of sharing vocabulary with the paired language in the case of Portuguese.


\begin{table}[h]
\centering
\small
\begin{tabular}{lccccc}
\toprule
& \multirow{2}{*}{\makecell{Vocab. \\size}} & \multicolumn{2}{c}{Kir-Eng} & \multicolumn{2}{c}{Kir-Por} \\
& & Kir & Eng & Kir & Por \\
\midrule
\multirow{3}{*}{\makecell{Combined \\Tokenisers}} & 10k & 1.52 &  1.48 & 1.51 & 1.61 \\
& 20k & 1.37 & 1.32 & 1.38 & 1.42 \\
& 30k & 1.29 & 1.25 & 1.29 & 1.32  \\
\midrule
\multirow{3}{*}{\makecell{Separate\\ Tokenisers}} & 10k & 1.51 & 1.55 & 1.51 & 1.72 \\
& 20k & 1.34 & 1.37 & 1.34 & 1.52  \\
& 30k & 1.29 & 1.29 & 1.29 & 1.40\\
\midrule
\multirow{3}{*}{\makecell{Difference in\\ tokeniser \\ fertility}} & 10k & 0.01 & -0.07 & 0.00 & -0.11 \\
& 20k & 0.03 & -0.05 & 0.04 & -0.10  \\
& 30k & 0.00 & -0.04 & 0.00 & -0.08 \\

\bottomrule
\end{tabular}
\caption{Fertility of combined Kiriol-English and Kiriol-Portuguese tokenisers and of separate Kiriol, English and Portuguese tokenisers on all training data. Negative differences indicate a tokeniser that has lower fertility (i.e. is more efficient) in the combined condition).}
\label{tab:fertility}
\end{table}


Based on prior work which shows that the encoding efficiency of a tokeniser depends on a language's morphological characteristics \citep{arnett2024language}, we explore whether these disparities are related to morphological differences between Portuguese, English and Kiriol. The Portuguese datasets have the most unique words and they are on average longer, while the Kiriol datasets have the least unique words with a shorter average word length (see \Cref{tab:lang_diffs}); this reflects that Portuguese has the greatest degree of inflectional morphology, and Kiriol the least \citep{koehn2005europarl}. The greater number of morphemes and word variability in the Portuguese and English data may result in more embedding space being allocated to Portuguese- and English-specific tokens than Kiriol-specific ones in a combined tokeniser setup, leading to the gains in tokeniser performance observed for Kiriol data when a separate tokeniser is used.

\begin{table}[h]
\centering
\small
\begin{tabular}{lcccc}
\toprule
& \multicolumn{2}{c}{Unstemmed} & \multicolumn{2}{c}{Stemmed} \\
Language & \makecell{\# unique \\words\\} & \makecell{Avg \\length} & \makecell{\# unique \\ words} & \makecell{Avg \\ length} \\
\midrule 
Kiriol & 50k & 4.50 & - & - \\
English & 66k & 4.85 & 30k & 4.12 \\ 
Portuguese & 91k & 5.12 & 29k & 3.77 \\ 
\bottomrule
\end{tabular}
\caption{Number of unique words and average lengths of words in all datasets in each language, across both regular and stemmed datasets.}
\label{tab:lang_diffs}
\end{table}

To test whether this hypothesis is correct, we explore whether the fertility of combined tokenisers on Kiriol data improves when the morphological complexity of the shared language (English or Portuguese) is synthetically reduced. We stem the Portuguese and English data, using the RSLP stemmer \citep{orengo2001stemming} and the English Porter stemmer \citep{porterstemmer} respectively, implemented in the NLTK toolkit \citep{loper2002nltknaturallanguagetoolkit}. This significantly reduces the number of unique morphemes present in each language's datasets (\Cref{tab:lang_diffs}). When we train combined tokenisers on the parallel data for Kiriol and stemmed English, and Kiriol and stemmed Portuguese, we indeed find that the fertility improves slightly for Kiriol across all vocabulary sizes (see \Cref{tab:fertility_stemmed}), even though the Kiriol data remains unchanged and unstemmed. This would indicate that, as the morphological complexity of the paired language is reduced, more vocabulary space is allocated towards tokens which are more useful for Kiriol data. 


\begin{table}[h]
\centering
\small
\begin{tabular}{ccccc}
\toprule
\makecell{Vocab. \\size} & 
Kir-Eng &
\makecell{Kir-Eng \\(Stemmed)} & 
Kir-Por &
\makecell{Kir-Por \\ (Stemmed)} \\
\midrule
10k & 1.52 & 1.49 & 1.51 & 1.47 \\
20k & 1.37 & 1.33 & 1.38 & 1.32 \\
30k & 1.29 & 1.28 & 1.29 & 1.28  \\
\bottomrule
\end{tabular}
\caption{Fertility of combined tokenisers for Kiriol-English and Kiriol-Portuguese, where the non-Kiriol language is either stemmed or unstemmed, on all Kiriol datasets.}
\label{tab:fertility_stemmed}
\end{table}

\subsubsection{Extrinsic Analysis}
\begin{figure}[h]
    \centering
    \includegraphics[width=1\linewidth]{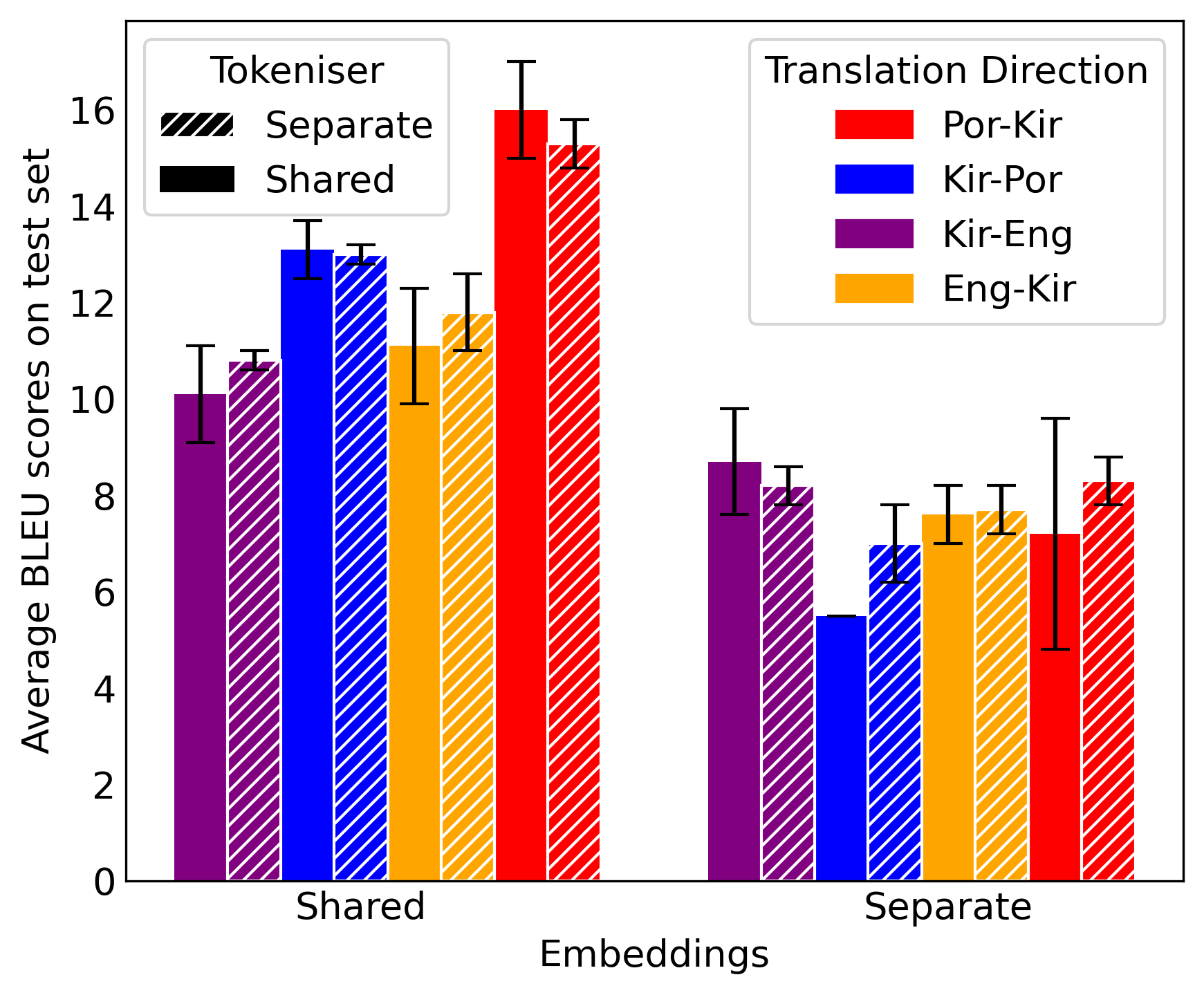}
    \caption{Average BLEU on test set using shared or separate embeddings, with combined (solid) and separate (hatched) tokenisers of vocabulary size 10k. Standard errors across model sets shown with error bars.}
    \label{fig:embeddings_10000}
\end{figure}

To investigate further the impacts of vocabulary sharing in both tokenisation and model training, we train new MT models on all of the training data, with combined and separate tokenisers, and shared and separate embedding spaces, for the three vocabulary sizes (10k, 20k, 30k). We show the average test scores for models trained with a vocabulary of 10k in \Cref{fig:embeddings_10000}, with full results and figures for the 20k and 30k vocabularies in \Cref{app:scores}.\footnote{For models trained with separate embeddings, there is high variability, with many models failing to train. To manage this, we remove any seeds which are more than 3.0 BLEU worse than the best-performing seed on the validation set.}

\begin{figure}[h]
\centering
\includegraphics[width=1\linewidth]{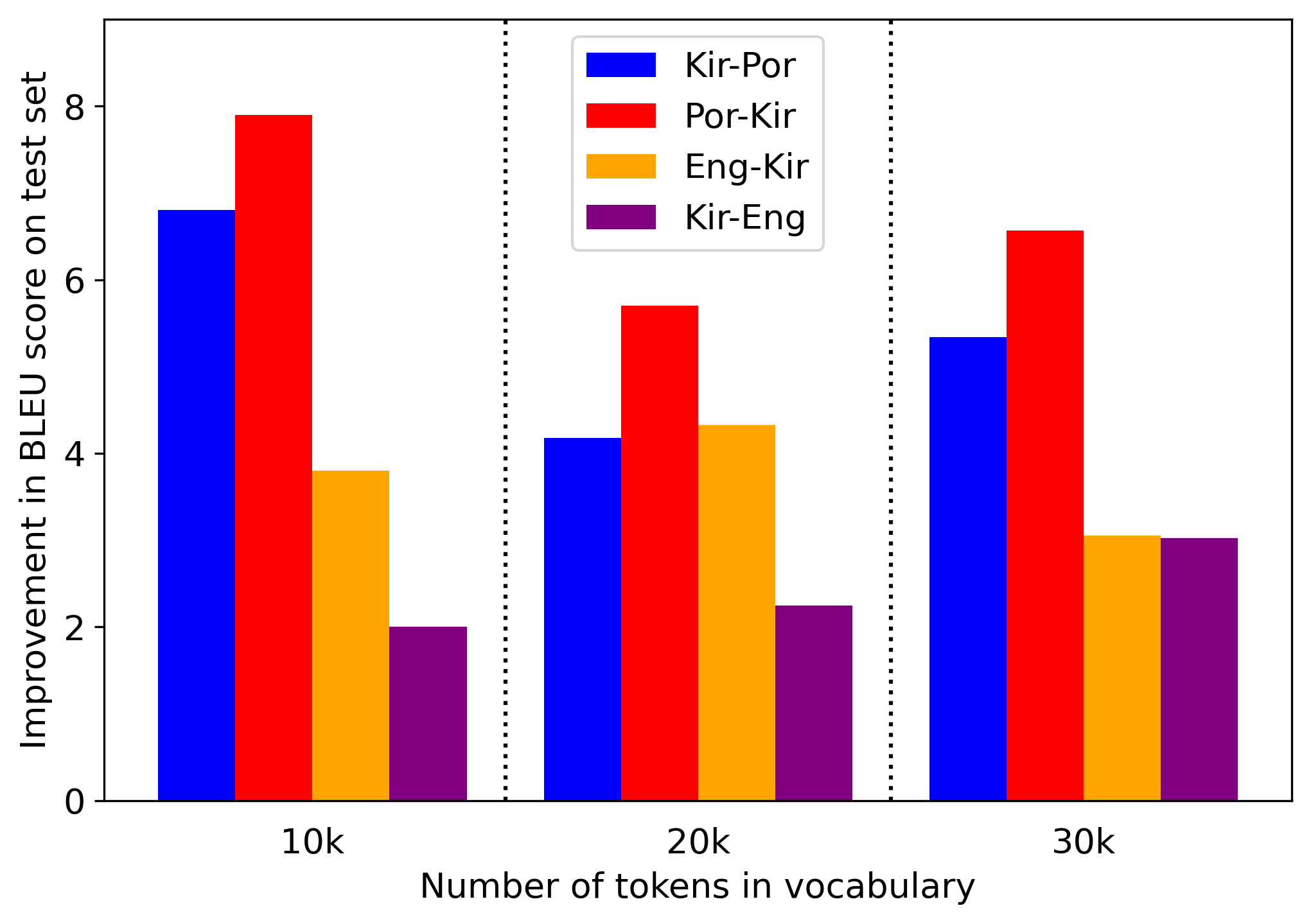}
\caption{Improvements in average BLEU on test set by models using shared embeddings compared to models using separate embeddings. Results averaged across combined and separate tokeniser conditions.}
\label{fig:change_bleu}
\end{figure}

While there is no consistent difference in models' BLEU scores across combined and separate tokeniser conditions, models with shared embeddings show better average performance than those with separate embeddings, across all language directions and vocabulary sizes. When we compare the size of the improvements in BLEU across the shared vs separate embeddings conditions (\Cref{fig:change_bleu}), Portuguese-Kiriol models show the greatest improvements, usually followed by the Kiriol-Portuguese models (except the 20k vocabulary setting). We tentatively take this as further evidence that lexical overlap between Kiriol and Portuguese is impacting model training, as a shared embedding space is more important for Portuguese/Kiriol models than English/Kiriol models. Further investigation is needed with more reliable optimisation and noise-reduction to corroborate this.

\section{Conclusion and Future Work}
We introduce a new dataset for the machine translation of Guinea-Bissau Creole, comprising around 40k sentences. This dataset is mostly from the Bible and Jehovah's Witnesses texts, with a small amount of general-domain data, and we use this to investigate how to optimise real-world translation performance with mostly religious text.  

While all of our translation models  show limited overall performance, our results provide insights into both the utility of different data types for domain generalisation and also the consistent differences in translation performance over different language pairs. While the performance of all of the models we train is limited, and human validation shows that overall results remain poor, we show that adding even a few hundred sentences of the target domain dictionary data to the Bible and WT training data considerably improved BLEU scores on the test set, indicating that this may be a strategy worth exploring in other low-resource scenarios. Collecting a small amount of manually translated sentences in the target domain should be possible in many low-resource MT settings where there is a specific domain of interest, and we show that these small sets can also be oversampled in the training data mix to further improve performance. 

We also investigate two well-known properties of creole languages, morphological simplicity and lexical overlap with a lexifier language. We show that differences in morphological complexity between Kiriol and English or Portuguese appear to result in differential tokenisation efficiency, whereby more tokenisation space is allocated to the more morphologically complex languages which disadvantages the more morphologically-simple Kiriol. We also show that using shared embeddings results in better average model performance over all languages, but that the difference is most pronounced for Kiriol-Portuguese and Portuguese-Kiriol models, indicating that the overlap in vocabulary between Kiriol and Portuguese can be better leveraged during training with shared embeddings. 

There are many avenues for future research even within these limited data scenarios, including exploring tokenisation strategies that specifically benefit creole-lexifier translation scenarios, or investigating how lexical overlap impacts other areas of model training outside of tokenisation and learning the embedding space; for example, by exploring patterns in attention mapping over input sentences. However, developing functional translation models for Kiriol will require either using Kiriol data to finetune pre-trained multilingual models, as in \citet{robinson-etal-2024-kreyol}, or collecting much more domain-diverse data from the Kiriol community to create better quality from-scratch models. As has been made clear elsewhere \citep{mager2023ethical, guridi2024supporting}, those research avenues will require careful consultation with the Kiriol-speaking community, to explore how they may want Kiriol to be represented—or not—in digital tools and technologies, and to find ways of making such research and language technology development maximally accessible to and beneficial for Kiriol speakers. 


\section*{Acknowledgments}
This work was partially supported by the UKRI AI Centre for Doctoral Training in Designing Responsible Natural Language Processing and the UKRI AI Centre for Doctoral Training in Speech and Language Technologies (SLT) and their Applications. Both centres are funded by UK Research and Innovation [grant numbers EP/S023062/1 and EP/Y030656/1 respectively]. For the purpose of open access, the authors have applied a Creative Commons Attribution (CC BY) licence to any Author Accepted Manuscript version arising.

\section*{Limitations}
\begin{compactitem}
\item Our investigations explore how the characteristics and relatedness of Kiriol and Portuguese impacts MT model training and performance, but these findings may not generalise to different creole-lexifier pairs. This is likely to depend on the way in which the creole in question formed, the degree of lexical overlap between the two languages, and the specific morphological structure and characteristics of the creole and lexifier languages in question. 
\item We rely on BLEU scores, an automated metric, as a proxy for translation quality, but these scores (particularly when they are so low) may not accurately reflect the quality of translations as well as other metrics or correlate with human judgements. 
\item Our work focuses on training from-scratch models, in order to investigate how fine-grained differences in training data and tokenisation setups impact downstream MT performance; but these findings may not scale to MT for Kiriol utilising finetuned models. 
\item As the bilingual dictionary was only parallel in Portuguese and Kiriol, we relied upon automatic translation to generate the English texts for this dataset. While each synthetic English sentence was manually checked against the Kiriol version by the lead author, who is fluent in both English and Kiriol, this method may impact the overall quality of the English translations. 
\end{compactitem}


\bibliography{acl_latex}

\begin{thebibliography}{42}
\providecommand{\natexlab}[1]{#1}

\bibitem[{Ahia and Ogueji(2020)}]{ahia2020towards}
Orevaoghene Ahia and Kelechi Ogueji. 2020.
\newblock \href {https://arxiv.org/abs/2003.12660} {Towards supervised and unsupervised neural machine translation baselines for nigerian pidgin}.
\newblock \emph{Preprint}, arXiv:2003.12660.

\bibitem[{Arnett and Bergen(2024)}]{arnett2024language}
Catherine Arnett and Benjamin~K. Bergen. 2024.
\newblock \href {https://arxiv.org/abs/2411.14198} {Why do language models perform worse for morphologically complex languages?}
\newblock \emph{Preprint}, arXiv:2411.14198.

\bibitem[{Birch et~al.(2008)Birch, Osborne, and Koehn}]{birch2008predicting}
Alexandra Birch, Miles Osborne, and Philipp Koehn. 2008.
\newblock \href {https://aclanthology.org/D08-1078/} {{Predicting Success in Machine Translation}}.
\newblock In \emph{Proceedings of the 2008 Conference on Empirical Methods in Natural Language Processing}, pages 745--754, Honolulu, Hawaii. Association for Computational Linguistics.

\bibitem[{Cotterell et~al.(2018)Cotterell, Mielke, Eisner, and Roark}]{cotterell2018all}
Ryan Cotterell, Sabrina~J. Mielke, Jason Eisner, and Brian Roark. 2018.
\newblock \href {https://doi.org/10.18653/v1/N18-2085} {{Are All Languages Equally Hard to Language-Model?}}
\newblock In \emph{Proceedings of the 2018 Conference of the North {A}merican Chapter of the Association for Computational Linguistics: Human Language Technologies, Volume 2 (Short Papers)}, pages 536--541, New Orleans, Louisiana. Association for Computational Linguistics.

\bibitem[{Dabre et~al.(2014)Dabre, Sukhoo, and Bhattacharyya}]{dabre-etal-2014-anou}
Raj Dabre, Aneerav Sukhoo, and Pushpak Bhattacharyya. 2014.
\newblock \href {https://aclanthology.org/W14-5113/} {{Anou Tradir: Experiences In Building Statistical Machine Translation Systems For Mauritian Languages {--} Creole, {E}nglish, {F}rench}}.
\newblock In \emph{Proceedings of the 11th International Conference on Natural Language Processing}, pages 82--88, Goa, India. NLP Association of India.

\bibitem[{Farquharson(2007)}]{farquharson2007creole}
Joseph Farquharson. 2007.
\newblock \href {https://doi.org/10.1075/tsl.73.04far} {{Creole Morphology Revisited}}.
\newblock In Umberto Ansaldo, Lisa Lim, and Stephen Matthews, editors, \emph{Deconstructing Creole}, pages 21--37. John Benjamins, Amsterdam.

\bibitem[{Gow-Smith and S{\'a}nchez~Villegas(2023)}]{gow-smith-snchez-villegas-2023-sheffields}
Edward Gow-Smith and Danae S{\'a}nchez~Villegas. 2023.
\newblock \href {https://doi.org/10.18653/v1/2023.americasnlp-1.21} {{{S}heffield`s Submission to the {A}mericas{NLP} Shared Task on Machine Translation into Indigenous Languages}}.
\newblock In \emph{Proceedings of the Workshop on Natural Language Processing for Indigenous Languages of the Americas (AmericasNLP)}, pages 192--199, Toronto, Canada. Association for Computational Linguistics.

\bibitem[{Guridi et~al.(2024)Guridi, Pertuze, and Zamora}]{guridi2024supporting}
Jose~A. Guridi, Julio~A. Pertuze, and Catalina Zamora. 2024.
\newblock \href {https://ceur-ws.org/Vol-3737/paper30.pdf} {{Supporting Participation Processes Using NLP in Constrained Resources Settings}}.
\newblock In \emph{Proceedings of the Ongoing Research, Practitioners, Posters, Workshops, and Projects of the International Conference EGOV-CeDEM-ePart 2024}, volume 3737, Belgium. Ghent University and KU Leuven.

\bibitem[{Haddow et~al.(2022)Haddow, Bawden, Miceli~Barone, Helcl, and Birch}]{haddow2022survey}
Barry Haddow, Rachel Bawden, Antonio~Valerio Miceli~Barone, Jind{\v{r}}ich Helcl, and Alexandra Birch. 2022.
\newblock \href {https://doi.org/10.1162/coli_a_00446} {{Survey of Low-Resource Machine Translation}}.
\newblock \emph{Computational Linguistics}, 48(3):673--732.

\bibitem[{Hutchinson(2024)}]{hutchinson2024modeling}
Ben Hutchinson. 2024.
\newblock \href {https://doi.org/10.18653/v1/2024.findings-naacl.65} {{Modeling the Sacred: Considerations when Using Religious Texts in Natural Language Processing}}.
\newblock In \emph{Findings of the Association for Computational Linguistics: NAACL 2024}, pages 1029--1043, Mexico City, Mexico. Association for Computational Linguistics.

\bibitem[{Kho et~al.(2024)Kho, Pham, Soon, and Chan}]{kho2024some}
Alastair Kho, Duc-Son Pham, Susannah Soon, and Kit~Yan Chan. 2024.
\newblock {Some Considerations for the Preservation of Endangered Languages Using Low-Resource Machine Translation}.
\newblock In \emph{Australasian Joint Conference on Artificial Intelligence}, pages 124--135. Springer Nature.

\bibitem[{Kingma and Ba(2015)}]{kingma2017adammethodstochasticoptimization}
Diederik~P. Kingma and Jimmy Ba. 2015.
\newblock \href {https://arxiv.org/abs/1412.6980} {Adam: A method for stochastic optimization}.
\newblock In \emph{3rd International Conference on Learning Representations, {ICLR} 2015, San Diego, CA, USA, May 7-9, 2015, Conference Track Proceedings}.

\bibitem[{Klein et~al.(2018)Klein, Kim, Deng, Nguyen, Senellart, and Rush}]{klein2018opennmt}
Guillaume Klein, Yoon Kim, Yuntian Deng, Vincent Nguyen, Jean Senellart, and Alexander Rush. 2018.
\newblock \href {https://aclanthology.org/W18-1817/} {{{O}pen{NMT}: Neural Machine Translation Toolkit}}.
\newblock In \emph{Proceedings of the 13th Conference of the Association for Machine Translation in the {A}mericas (Volume 1: Research Track)}, pages 177--184, Boston, MA. Association for Machine Translation in the Americas.

\bibitem[{Koehn(2005)}]{koehn2005europarl}
Philipp Koehn. 2005.
\newblock \href {https://aclanthology.org/2005.mtsummit-papers.11/} {{{E}uroparl: A Parallel Corpus for Statistical Machine Translation}}.
\newblock In \emph{Proceedings of Machine Translation Summit X: Papers}, pages 79--86, Phuket, Thailand.

\bibitem[{Koehn and Monz(2006)}]{koehn-monz-2006-manual}
Philipp Koehn and Christof Monz. 2006.
\newblock \href {https://aclanthology.org/W06-3114/} {Manual and automatic evaluation of machine translation between {E}uropean languages}.
\newblock In \emph{Proceedings on the Workshop on Statistical Machine Translation}, pages 102--121, New York City. Association for Computational Linguistics.

\bibitem[{Kohl(2016)}]{kohl2016limitations}
Christoph Kohl. 2016.
\newblock {Limitations and Ambiguities of Colonialism in Guinea-Bissau: Examining the Creole and “Civilized” Space in Colonial Society}.
\newblock \emph{History in Africa}, 43:169--203.

\bibitem[{Kohl(2018)}]{kohl2018creole}
Christoph Kohl. 2018.
\newblock Creole language and identity in guinea-bissau: Socio-anthropological perspectives.
\newblock In \emph{Creolization and Pidginization in Contexts of Postcolonial Diversity}, pages 158--177. Brill.

\bibitem[{Kudo(2018)}]{kudo2018subword}
Taku Kudo. 2018.
\newblock \href {https://doi.org/10.18653/v1/P18-1007} {{Subword Regularization: Improving Neural Network Translation Models with Multiple Subword Candidates}}.
\newblock In \emph{Proceedings of the 56th Annual Meeting of the Association for Computational Linguistics (Volume 1: Long Papers)}, pages 66--75, Melbourne, Australia. Association for Computational Linguistics.

\bibitem[{Lent et~al.(2021)Lent, Bugliarello, de~Lhoneux, Qiu, and S{\o}gaard}]{lent-etal-2021-language}
Heather Lent, Emanuele Bugliarello, Miryam de~Lhoneux, Chen Qiu, and Anders S{\o}gaard. 2021.
\newblock \href {https://doi.org/10.18653/v1/2021.conll-1.5} {{On Language Models for Creoles}}.
\newblock In \emph{Proceedings of the 25th Conference on Computational Natural Language Learning}, pages 58--71, Online. Association for Computational Linguistics.

\bibitem[{Lent et~al.(2024)Lent, Tatariya, Dabre, Chen, Fekete, Ploeger, Zhou, Armstrong, Eijansantos, Malau, Heje, Lavrinovics, Kanojia, Belony, Bollmann, Grobol, Lhoneux, Hershcovich, DeGraff, S{\o}gaard, and Bjerva}]{lent2024creoleval}
Heather Lent, Kushal Tatariya, Raj Dabre, Yiyi Chen, Marcell Fekete, Esther Ploeger, Li~Zhou, Ruth-Ann Armstrong, Abee Eijansantos, Catriona Malau, Hans~Erik Heje, Ernests Lavrinovics, Diptesh Kanojia, Paul Belony, Marcel Bollmann, Lo{\"i}c Grobol, Miryam~de Lhoneux, Daniel Hershcovich, Michel DeGraff, Anders S{\o}gaard, and Johannes Bjerva. 2024.
\newblock \href {https://doi.org/10.1162/tacl_a_00682} {{{C}reole{V}al: Multilingual Multitask Benchmarks for Creoles}}.
\newblock \emph{Transactions of the Association for Computational Linguistics}, 12:950--978.

\bibitem[{Liu et~al.(2021)Liu, Ryan, and Hulden}]{liu-etal-2021-usefulness}
Ling Liu, Zach Ryan, and Mans Hulden. 2021.
\newblock \href {https://aclanthology.org/2021.computel-1.6/} {{The Usefulness of {B}ibles in Low-Resource Machine Translation}}.
\newblock In \emph{Proceedings of the 4th Workshop on the Use of Computational Methods in the Study of Endangered Languages Volume 1 (Papers)}, pages 44--50, Online. Association for Computational Linguistics.

\bibitem[{Loper and Bird(2002)}]{loper2002nltknaturallanguagetoolkit}
Edward Loper and Steven Bird. 2002.
\newblock \href {https://doi.org/10.3115/1118108.1118117} {{{NLTK}: The Natural Language Toolkit}}.
\newblock In \emph{Proceedings of the {ACL}-02 Workshop on Effective Tools and Methodologies for Teaching Natural Language Processing and Computational Linguistics}, pages 63--70, Philadelphia, Pennsylvania, USA. Association for Computational Linguistics.

\bibitem[{Mager et~al.(2023)Mager, Mager, Kann, and Vu}]{mager2023ethical}
Manuel Mager, Elisabeth Mager, Katharina Kann, and Ngoc~Thang Vu. 2023.
\newblock \href {https://doi.org/10.18653/v1/2023.acl-long.268} {{Ethical Considerations for Machine Translation of Indigenous Languages: Giving a Voice to the Speakers}}.
\newblock In \emph{Proceedings of the 61st Annual Meeting of the Association for Computational Linguistics (Volume 1: Long Papers)}, pages 4871--4897, Toronto, Canada. Association for Computational Linguistics.

\bibitem[{Marashian et~al.(2025)Marashian, Rice, Gessler, Palmer, and von~der Wense}]{marashian-etal-2025-priest}
Ali Marashian, Enora Rice, Luke Gessler, Alexis Palmer, and Katharina von~der Wense. 2025.
\newblock \href {https://aclanthology.org/2025.coling-main.472/} {{From Priest to Doctor: Domain Adaptation for Low-Resource Neural Machine Translation}}.
\newblock In \emph{Proceedings of the 31st International Conference on Computational Linguistics}, pages 7087--7098, Abu Dhabi, UAE. Association for Computational Linguistics.

\bibitem[{Moreira and Huyck(2001)}]{orengo2001stemming}
Viviane Moreira and Chris Huyck. 2001.
\newblock \href {https://doi.org/10.1109/SPIRE.2001.989755} {{A Stemming Algorithmm for the Portuguese Language.}}
\newblock In \emph{Proceedings of the Eighth Symposium on String Processing and Information Retrieval}, pages 186--193.

\bibitem[{Mueller et~al.(2020)Mueller, Nicolai, McCarthy, Lewis, Wu, and Yarowsky}]{mueller-etal-2020-analysis}
Aaron Mueller, Garrett Nicolai, Arya~D. McCarthy, Dylan Lewis, Winston Wu, and David Yarowsky. 2020.
\newblock \href {https://aclanthology.org/2020.lrec-1.458/} {An analysis of massively multilingual neural machine translation for low-resource languages}.
\newblock In \emph{Proceedings of the Twelfth Language Resources and Evaluation Conference}, pages 3710--3718, Marseille, France. European Language Resources Association.

\bibitem[{Mufwene(1996)}]{mufwene1996founder}
Salikoko~S Mufwene. 1996.
\newblock {The founder principle in creole genesis}.
\newblock \emph{Diachronica}, 13(1):83--134.

\bibitem[{NLLB(2024)}]{nllb2024scaling}
Team NLLB. 2024.
\newblock Scaling neural machine translation to 200 languages.
\newblock \emph{Nature}, 630(8018):841--846.

\bibitem[{Papineni et~al.(2002)Papineni, Roukos, Ward, and Zhu}]{papineni2002bleu}
Kishore Papineni, Salim Roukos, Todd Ward, and Wei-Jing Zhu. 2002.
\newblock \href {https://doi.org/10.3115/1073083.1073135} {{{B}leu: a Method for Automatic Evaluation of Machine Translation}}.
\newblock In \emph{Proceedings of the 40th Annual Meeting of the Association for Computational Linguistics}, pages 311--318, Philadelphia, Pennsylvania, USA. Association for Computational Linguistics.

\bibitem[{Park et~al.(2021)Park, Zhang, Haley, Steimel, Liu, and Schwartz}]{park2021morphology}
Hyunji~Hayley Park, Katherine~J. Zhang, Coleman Haley, Kenneth Steimel, Han Liu, and Lane Schwartz. 2021.
\newblock \href {https://doi.org/10.1162/tacl_a_00365} {{Morphology Matters: A Multilingual Language Modeling Analysis}}.
\newblock \emph{Transactions of the Association for Computational Linguistics}, 9:261--276.

\bibitem[{Porter and Boulton(2001)}]{porterstemmer}
Martin Porter and Richard Boulton. 2001.
\newblock \href {http://snowball.tartarus.org/algorithms/english/stemmer.html} {\emph{{The English (Porter2) stemming algorithm}}}.

\bibitem[{Post(2018)}]{post-2018-call}
Matt Post. 2018.
\newblock \href {https://doi.org/10.18653/v1/W18-6319} {{A Call for Clarity in Reporting {BLEU} Scores}}.
\newblock In \emph{Proceedings of the Third Conference on Machine Translation: Research Papers}, pages 186--191, Brussels, Belgium. Association for Computational Linguistics.

\bibitem[{Robinson et~al.(2024)Robinson, Dabre, Shurtz, Dent, Onesi, Monroc, Grobol, Muhammad, Garg, Etori, Tiyyala, Samuel, Stutzman, Odoom, Khudanpur, Richardson, and Murray}]{robinson-etal-2024-kreyol}
Nathaniel Robinson, Raj Dabre, Ammon Shurtz, Rasul Dent, Onenamiyi Onesi, Claire Monroc, Lo{\"i}c Grobol, Hasan Muhammad, Ashi Garg, Naome Etori, Vijay~Murari Tiyyala, Olanrewaju Samuel, Matthew Stutzman, Bismarck Odoom, Sanjeev Khudanpur, Stephen Richardson, and Kenton Murray. 2024.
\newblock \href {https://doi.org/10.18653/v1/2024.naacl-long.170} {{Krey{\`o}l-{MT}: Building {MT} for {L}atin {A}merican, {C}aribbean and Colonial {A}frican Creole Languages}}.
\newblock In \emph{Proceedings of the 2024 Conference of the North American Chapter of the Association for Computational Linguistics: Human Language Technologies (Volume 1: Long Papers)}, pages 3083--3110, Mexico City, Mexico. Association for Computational Linguistics.

\bibitem[{Robinson et~al.(2022)Robinson, Hogan, Fulda, and Mortensen}]{robinson-etal-2022-data}
Nathaniel Robinson, Cameron Hogan, Nancy Fulda, and David~R. Mortensen. 2022.
\newblock \href {https://aclanthology.org/2022.loresmt-1.5/} {{Data-adaptive Transfer Learning for Translation: A Case Study in {H}aitian and Jamaican}}.
\newblock In \emph{Proceedings of the Fifth Workshop on Technologies for Machine Translation of Low-Resource Languages (LoResMT 2022)}, pages 35--42, Gyeongju, Republic of Korea. Association for Computational Linguistics.

\bibitem[{Rust et~al.(2021)Rust, Pfeiffer, Vuli{\'c}, Ruder, and Gurevych}]{rust-etal-2021-good}
Phillip Rust, Jonas Pfeiffer, Ivan Vuli{\'c}, Sebastian Ruder, and Iryna Gurevych. 2021.
\newblock \href {https://doi.org/10.18653/v1/2021.acl-long.243} {{How Good is Your Tokenizer? On the Monolingual Performance of Multilingual Language Models}}.
\newblock In \emph{Proceedings of the 59th Annual Meeting of the Association for Computational Linguistics and the 11th International Joint Conference on Natural Language Processing (Volume 1: Long Papers)}, pages 3118--3135, Online. Association for Computational Linguistics.

\bibitem[{Sennrich et~al.(2016)Sennrich, Haddow, and Birch}]{sennrich2015neural}
Rico Sennrich, Barry Haddow, and Alexandra Birch. 2016.
\newblock \href {https://doi.org/10.18653/v1/P16-1162} {{Neural Machine Translation of Rare Words with Subword Units}}.
\newblock In \emph{Proceedings of the 54th Annual Meeting of the Association for Computational Linguistics (Volume 1: Long Papers)}, pages 1715--1725, Berlin, Germany. Association for Computational Linguistics.

\bibitem[{Siddhant et~al.(2022)Siddhant, Bapna, Firat, Cao, Chen, Caswell, and Garcia}]{siddhant2022towards}
Aditya Siddhant, Ankur Bapna, Orhan Firat, Yuan Cao, Mia~Xu Chen, Isaac Caswell, and Xavier Garcia. 2022.
\newblock \href {https://arxiv.org/abs/2201.03110} {{Towards the Next 1000 Languages in Multilingual Machine Translation: Exploring the Synergy Between Supervised and Self-Supervised Learning}}.
\newblock \emph{CoRR}, abs/2201.03110.

\bibitem[{Siegel(1999)}]{siegel1999creoles}
Jeff Siegel. 1999.
\newblock Creoles and minority dialects in education: An overview.
\newblock \emph{Journal of Multilingual and Multicultural Development}, 20(6):508--531.

\bibitem[{Vaswani et~al.(2017)Vaswani, Shazeer, Parmar, Uszkoreit, Jones, Gomez, Kaiser, and Polosukhin}]{vaswani2017attention}
Ashish Vaswani, Noam Shazeer, Niki Parmar, Jakob Uszkoreit, Llion Jones, Aidan~N. Gomez, \L{}ukasz Kaiser, and Illia Polosukhin. 2017.
\newblock \href {https://proceedings.neurips.cc/paper_files/paper/2017/file/3f5ee243547dee91fbd053c1c4a845aa-Paper.pdf} {Attention is all you need}.
\newblock In \emph{Proceedings of the 31st International Conference on Neural Information Processing Systems}, NIPS'17, page 6000–6010, Red Hook, NY, USA. Curran Associates Inc.

\bibitem[{V{\'a}zquez et~al.(2021)V{\'a}zquez, Scherrer, Virpioja, and Tiedemann}]{vazquez-etal-2021-helsinki}
Ra{\'u}l V{\'a}zquez, Yves Scherrer, Sami Virpioja, and J{\"o}rg Tiedemann. 2021.
\newblock \href {https://doi.org/10.18653/v1/2021.americasnlp-1.29} {The {H}elsinki submission to the {A}mericas{NLP} shared task}.
\newblock In \emph{Proceedings of the First Workshop on Natural Language Processing for Indigenous Languages of the Americas}, pages 255--264, Online. Association for Computational Linguistics.

\bibitem[{Wigglesworth et~al.(2013)Wigglesworth, Billington, and Loakes}]{wigglesworth2013creole}
Gillian Wigglesworth, Rosey Billington, and Deborah Loakes. 2013.
\newblock \href {https://doi.org/10.1111/lnc3.12035} {Creole speakers and standard language education}.
\newblock \emph{Linguistics and Language Compass}, 7(7):388--397.

\bibitem[{Zwennicker and Stap(2022)}]{zwennicker2022towards}
Just Zwennicker and David Stap. 2022.
\newblock \href {https://www.winlp.org/wp-content/uploads/2022/11/35_Paper.pdf} {Towards a general purpose machine translation system for sranantongo}.
\newblock In \emph{Proceedings of the 2022 EMNLP Workshop WiNLP}, Abu Dhabi, United Arab Emirates (Hybrid).

\end{thebibliography}

\clearpage

\appendix

\section{Parameters}
\label{app:params}

In \Cref{tab:model_parameters}, we provide a full list of model parameters:
\begin{table}[h]
    \centering
    \begin{tabular}{l l}
        \toprule
        \textbf{Model Parameters} & \textbf{Value} \\
        \midrule
        Hidden size & 512 \\
        Layers & 6 \\
        Heads & 8 \\
        Dropout & 0.1 \\
        Attention dropout & 0.1 \\
        Train steps & 20k \\
        Warmup steps & 1,000 \\
        Validation steps & 1,000 \\
        Validation metrics & BLEU \\
        Early stopping & 5 \\
        Src words min frequency & 2 \\
        Tgt words min frequency & 2 \\
        Batch type & Tokens \\
        Batch size & 4,096 \\
        Validation batch size & 2,048 \\
        Accum count & 4 \\
        Accum steps & 0 \\
        Compute dtype & FP16 \\
        Optimizer & Adam \\
        Learning rate & 2 \\
        Warmup steps & 1,000 \\
        Decay method & Noam \\
        Adam beta2 & 0.998 \\
        \midrule
        \textbf{Inference} & \\
        Beam size & 5 \\
        \bottomrule
    \end{tabular}
    \caption{Model Parameters}
    \label{tab:model_parameters}
\end{table}

\section{Average model performance scores and standard errors}
\label{app:scores}

\begin{figure}[h]
\centering
\includegraphics[width=1\linewidth]{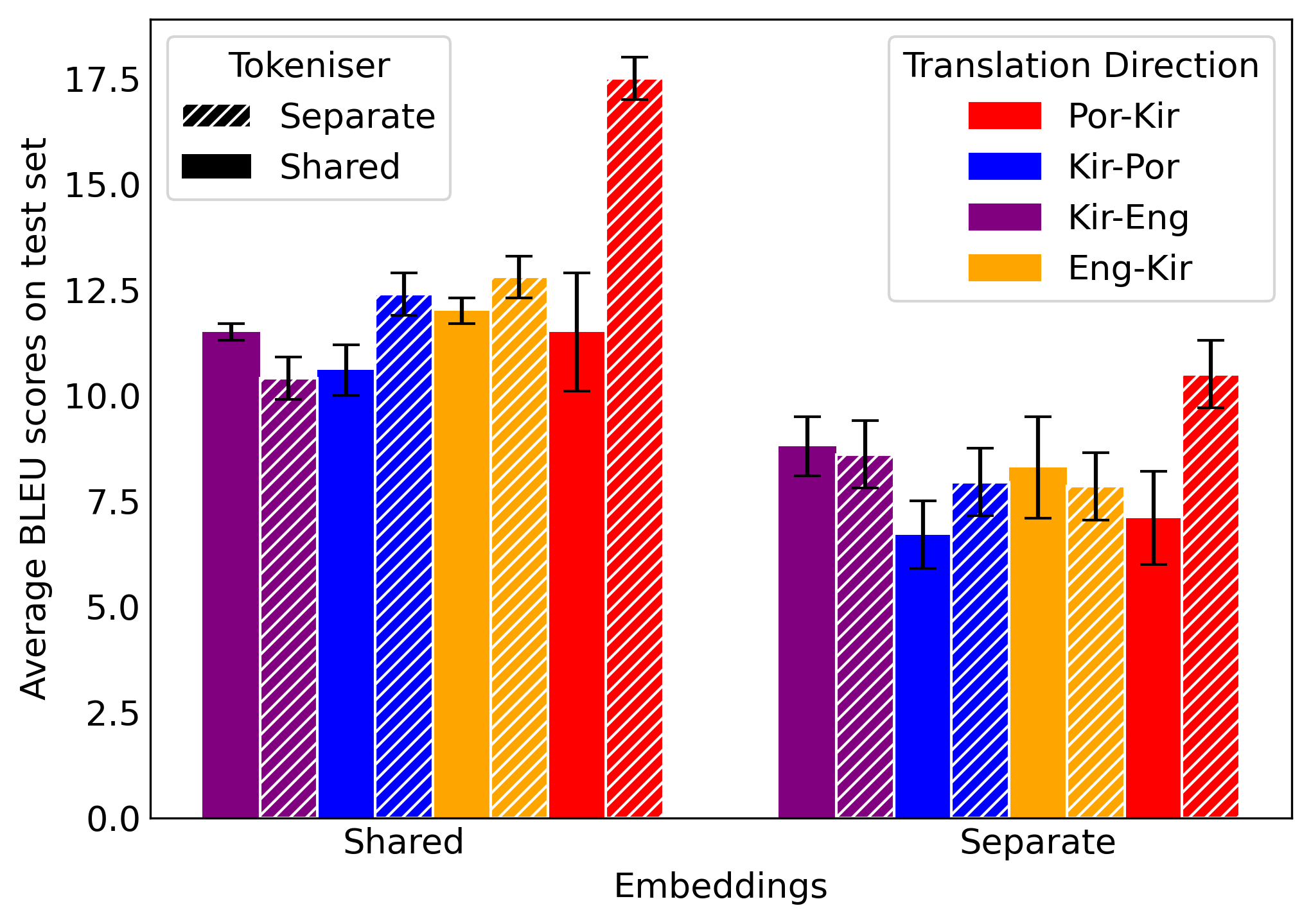}
\caption{Average BLEU scores on test set using shared or separate embeddings, with combined (solid) and separate (hatched) tokenisers of vocabulary size 20k. Standard errors across model sets are shown with error bars.}
\label{fig:embeddings_20000}
\end{figure}

\begin{figure}[h]
\centering
\includegraphics[width=1\linewidth]{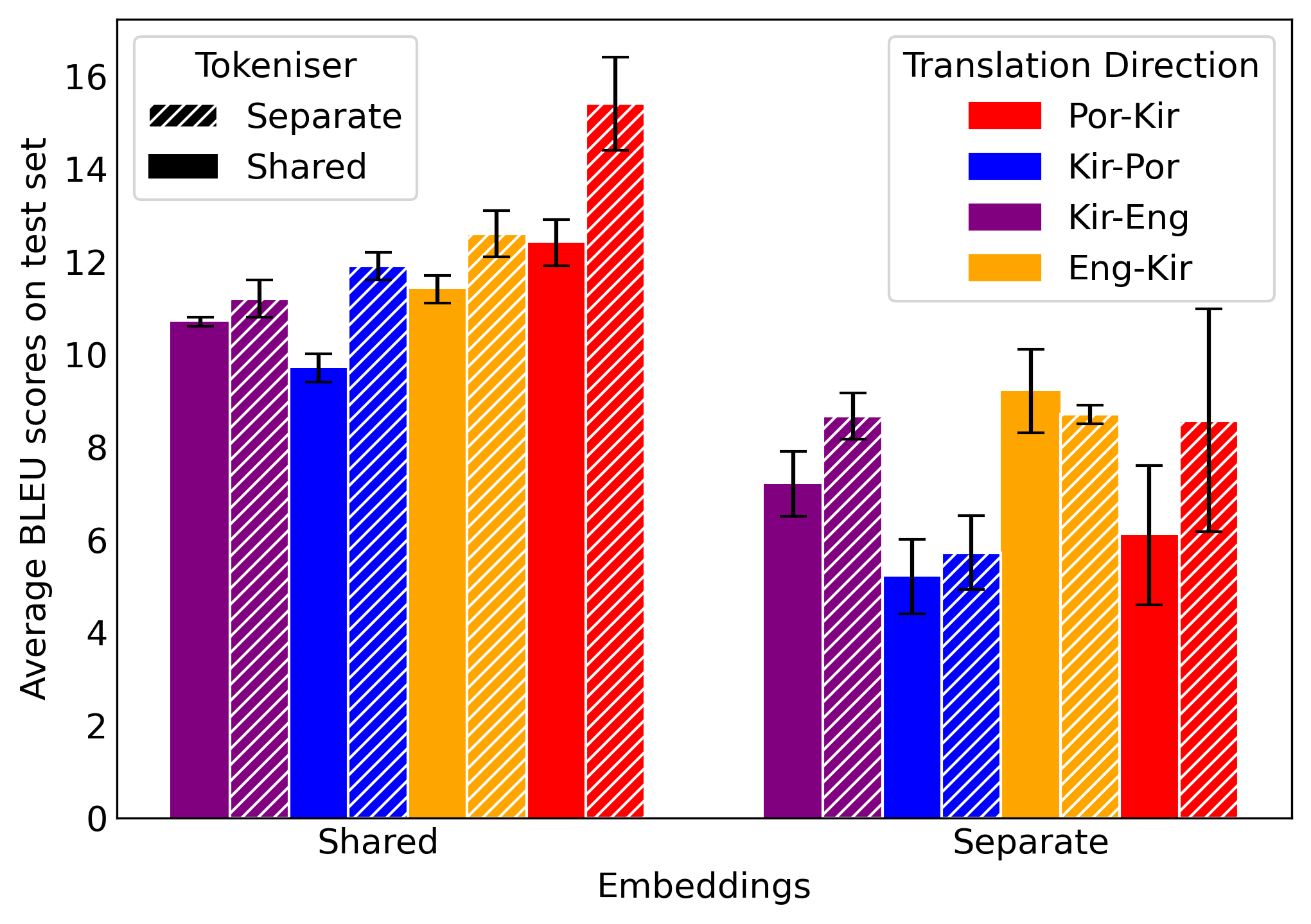}
\caption{Average BLEU scores on test set using shared or separate embeddings, with combined (solid) and separate (hatched) tokenisers of vocabulary size 30k. Standard errors across model sets are shown with error bars.}
\label{fig:embeddings_30000}
\end{figure}

In \Cref{tab:full_results_1}, we show the average scores across 5 model seeds for each training data setup and language direction from \Cref{section:4.1}. In \Cref{tab:full_results_2} we do the same for models from \Cref{section:4.2}. In \Cref{tab:full_results_3} we show the results of all models trained for the experiments in \Cref{section:4.3}, and those with the anomalous seed models removed as reported above in \Cref{tab:full_results_4}. \Cref{fig:embeddings_20000} and \Cref{fig:embeddings_30000} show the corresponding results for \Cref{fig:embeddings_10000} for vocabulary sizes of 20,000 and 30,000 respectively. 

\begin{table*}
\resizebox{\linewidth}{!}{

\begin{tabular}{llllllllllllllllllll}
\toprule
& & \multicolumn{2}{c}{Bible} & \multicolumn{2}{c}{WT} & \multicolumn{2}{c}{Bible \& WT} & \multicolumn{2}{c}{NT \& WT} & \multicolumn{2}{c}{OT \& WT} & \multicolumn{2}{c}{\makecell{Bible \& WT \\(shortest 50\%)}} & \multicolumn{2}{c}{\makecell{Bible \& WT \\(longest 50\%)}} & \multicolumn{2}{c}{\makecell{Bible \& WT \\(undersampled)}} & \multicolumn{2}{c}{\makecell{Bible \& WT \\(oversampled)}} \\
\vspace{5pt}
& & \multicolumn{2}{c}{(30k)} & \multicolumn{2}{c}{(7k)} & \multicolumn{2}{c}{(36k)}& \multicolumn{2}{c}{(14k)} & \multicolumn{2}{c}{(29k)} & \multicolumn{2}{c}{(18k)} & \multicolumn{2}{c}{(18k)} & \multicolumn{2}{c}{(13k)} & \multicolumn{2}{c}{(59k)} \\
 Language Pair & Evaluation Set  & Avg & StdE & Avg & StdE & Avg & StdE & Avg & StdE & Avg & StdE & Avg & StdE & Avg & StdE & Avg & StdE & Avg & StdE \\
\midrule
\multirow[c]{4}{*}{Kir-Eng} & Val bible & 20.88 & 0.95 & 0.78 & 0.41 & 19.74 & 0.72 & 6.64 & 1.18 & 17.88 & 0.62 & 14.26 & 0.65 & 13.68 & 1.96 & 10.40 & 1.24 & 18.30 & 0.72 \\

 & Val WT & 0.98 & 0.15 & 4.26 & 1.77 & 8.84 & 0.20 & 7.14 & 1.32 & 8.90 & 0.27 & 1.86 & 0.08 & 7.00 & 1.25 & 7.86 & 1.26 & 9.74 & 0.14 \\
 
 & Val all & 10.93 & 0.44 & 2.52 & 1.08 & 14.29 & 0.43 & 6.89 & 1.24 & 13.39 & 0.39 & 8.06 & 0.34 & 10.34 & 1.60 & 9.13 & 1.25 & 14.02 & 0.36 \\
 
 & Test & 3.92 & 0.29 & 0.58 & 0.35 & 4.54 & 0.21 & 2.00 & 0.62 & 3.64 & 0.25 & 3.66 & 0.18 & 0.62 & 0.17 & 1.90 & 0.49 & 4.28 & 0.23 \\
 
\cline{1-20}

\multirow[c]{4}{*}{Kir-Por} & Val bible & 19.50 & 1.43 & 0.18 & 0.07 & 18.94 & 0.37 & 7.84 & 0.54 & 18.20 & 0.23 & 14.00 & 0.30 & 14.30 & 2.52 & 9.22 & 0.74 & 9.30 & 3.66 \\

 & Val WT & 0.42 & 0.04 & 2.50 & 0.81 & 7.58 & 0.32 & 7.84 & 0.28 & 7.98 & 0.14 & 2.06 & 0.05 & 6.22 & 1.13 & 6.06 & 0.66 & 4.62 & 1.49 \\
 
 & Val all & 9.96 & 0.73 & 1.34 & 0.44 & 13.26 & 0.30 & 7.84 & 0.40 & 13.09 & 0.18 & 8.03 & 0.14 & 10.26 & 1.82 & 7.64 & 0.69 & 6.96 & 2.57 \\
 
 & Test & 2.40 & 0.16 & 0.10 & 0.03 & 3.12 & 0.46 & 2.68 & 0.08 & 3.38 & 0.15 & 2.44 & 0.11 & 0.48 & 0.22 & 1.56 & 0.36 & 1.50 & 0.69 \\
 
\cline{1-20}
\multirow[c]{4}{*}{Eng-Kir} & Val bible & 20.62 & 1.62 & 0.30 & 0.05 & 22.52 & 1.52 & 9.06 & 0.71 & 16.58 & 2.63 & 13.12 & 1.48 & 14.90 & 2.76 & 9.72 & 1.86 & 15.92 & 3.08 \\

 & Val WT & 1.48 & 0.10 & 2.52 & 0.37 & 8.24 & 0.63 & 8.36 & 0.60 & 7.50 & 1.06 & 2.36 & 0.32 & 5.94 & 1.23 & 6.48 & 1.17 & 7.44 & 1.26 \\
 
 & Val all & 11.05 & 0.78 & 1.41 & 0.21 & 15.38 & 1.07 & 8.71 & 0.65 & 12.04 & 1.83 & 7.74 & 0.89 & 10.42 & 1.99 & 8.10 & 1.51 & 11.68 & 2.14 \\
 
 & Test & 3.92 & 0.28 & 0.08 & 0.02 & 4.12 & 0.65 & 2.52 & 0.42 & 3.82 & 0.84 & 2.78 & 0.60 & 1.66 & 0.40 & 1.80 & 0.48 & 3.72 & 0.84 \\
\cline{1-20}

\multirow[c]{4}{*}{Por-Kir} & Val bible & 23.64 & 0.75 & 0.14 & 0.06 & 21.06 & 2.83 & 7.32 & 1.28 & 19.82 & 0.98 & 15.26 & 0.44 & 15.70 & 1.78 & 8.76 & 2.51 & 17.26 & 3.56 \\

 & Val WT & 1.18 & 0.06 & 0.96 & 0.43 & 8.12 & 0.99 & 6.68 & 0.92 & 8.26 & 0.48 & 2.22 & 0.13 & 6.16 & 0.85 & 5.52 & 1.48 & 7.50 & 1.33 \\
 
 & Val all & 12.41 & 0.39 & 0.55 & 0.24 & 14.59 & 1.91 & 7.00 & 1.10 & 14.04 & 0.72 & 8.74 & 0.28 & 10.93 & 1.31 & 7.14 & 2.00 & 12.38 & 2.43 \\
 
 & Test & 4.58 & 0.40 & 0.04 & 0.02 & 5.14 & 1.10 & 2.22 & 0.72 & 4.16 & 0.60 & 4.12 & 0.04 & 2.00 & 0.54 & 1.80 & 0.95 & 4.64 & 1.10 \\
\bottomrule
\end{tabular}
}
\caption{Average performance and standard errors for all sets of models from Section 5.1 on the test set of 1,603 domain-general dictionary sentences and the validation datasets of 500 sentences from the Bible and 500 from the WT. Each is averaged over five random seed models. Number of sentences in each training data setup shown in brackets.}
\label{tab:full_results_1}
\end{table*}

\begin{table*}
\resizebox{\linewidth}{!}{
\centering
\tiny
\begin{tabular}{llllllllllllllll}
\toprule
& & \multicolumn{2}{c}{BWT \& Lexicon} & \multicolumn{2}{c}{BWT \& Donations} & \multicolumn{2}{c}{\makecell{BWT \& 300 Dict. \\sentences}} & \multicolumn{2}{c}{\makecell{BWT \& 600 Dict. \\sentences}} & \multicolumn{2}{c}{\makecell{BWT \& 600 Dict. \\sentences x 5}} & \multicolumn{2}{c}{\makecell{BWT \& 600 Dict. \\sentences x 10}} & \multicolumn{2}{c}{BTW \& all} \\
\vspace{5pt}
& & \multicolumn{2}{c}{(37.7k)} & \multicolumn{2}{c}{(36.0k)} & \multicolumn{2}{c}{(36.1k)}& \multicolumn{2}{c}{(36.4k)} & \multicolumn{2}{c}{(38.8k)} & \multicolumn{2}{c}{(41.8k)} & \multicolumn{2}{c}{(41.0k)} \\
Language Pair & Evaluation Set  & Avg & StdE & Avg & StdE & Avg & StdE & Avg & StdE & Avg & StdE & Avg & StdE & Avg & StdE \\
\midrule

\multirow[c]{4}{*}{Kir-Eng} & Val bible & 18.84 & 0.78 & 19.34 & 0.71 & 19.80 & 0.11 & 20.18 & 0.53 & 17.74 & 3.90 & 17.40 & 2.96 & 19.68 & 0.52 \\

 & Val WT & 8.60 & 0.23 & 8.66 & 0.28 & 9.28 & 0.23 & 9.30 & 0.15 & 8.06 & 1.62 & 8.22 & 1.35 & 9.08 & 0.37 \\
 
 & Val all & 13.72 & 0.44 & 14.00 & 0.48 & 14.54 & 0.16 & 14.74 & 0.32 & 12.90 & 2.76 & 12.81 & 2.15 & 14.38 & 0.42 \\
 
 & Test & 4.58 & 0.44 & 4.06 & 0.22 & 7.56 & 0.20 & 9.48 & 0.46 & 8.88 & 2.10 & 8.24 & 1.73 & 10.14 & 0.96 \\
 
\cline{1-16}
\multirow[c]{4}{*}{Kir-Por} & Val bible & 20.7 & 0.44 & 20.04 & 0.79 & 19.88 & 0.45 & 19.94 & 0.34 & 20.46 & 0.36 & 19.86 & 0.49 & 19.16 & 0.96 \\

 & Val WT & 7.98 & 0.10 & 7.90 & 0.05 & 8.00 & 0.08 & 7.88 & 0.15 & 7.80 & 0.13 & 7.50 & 0.15 & 8.20 & 0.13 \\
 
 & Val all & 14.34 & 0.19 & 13.97 & 0.41 & 13.94 & 0.24 & 13.91 & 0.22 & 14.13 & 0.18 & 13.68 & 0.27 & 13.68 & 0.41 \\
 
 & Test & 4.10 & 0.14 & 4.04 & 0.15 & 7.88 & 0.31 & 10.22 & 0.41 & 12.66 & 0.26 & 10.32 & 0.84 & 13.06 & 0.60 \\
 
\cline{1-16}
\multirow[c]{4}{*}{Eng-Kir} & Val bible & 22.60 & 0.56 & 21.36 & 1.72 & 22.04 & 0.59 & 22.88 & 0.71 & 22.58 & 0.87 & 22.80 & 0.48 & 22.34 & 0.65 \\

 & Val WT & 9.06 & 0.12 & 7.92 & 0.84 & 8.54 & 0.30 & 9.08 & 0.25 & 9.18 & 0.12 & 10.64 & 0.16 & 9.02 & 0.17 \\
 
 & Val all & 15.83 & 0.31 & 14.64 & 1.27 & 15.29 & 0.43 & 15.98 & 0.47 & 15.88 & 0.44 & 16.72 & 0.32 & 15.68 & 0.36 \\
 
 & Test & 4.90 & 0.38 & 3.7- & 0.42 & 7.26 & 0.24 & 8.90 & 0.47 & 11.00 & 0.31 & 9.84 & 0.74 & 11.14 & 1.17 \\
 
\cline{1-16}
\multirow[c]{4}{*}{Por-Kir} & Val bible & 23.58 & 0.37 & 23.04 & 0.69 & 23.82 & 0.41 & 20.16 & 3.73 & 19.34 & 3.85 & 23.04 & 0.67 & 19.24 & 4.04 \\

 & Val WT & 9.16 & 0.05 & 9.42 & 0.26 & 9.34 & 0.10 & 7.84 & 1.39 & 7.66 & 1.40 & 8.72 & 0.42 & 7.76 & 1.45 \\
 
 & Val all & 16.37 & 0.19 & 16.23 & 0.46 & 16.58 & 0.25 & 14.00 & 2.56 & 13.5 & 2.62 & 15.88 & 0.55 & 13.5 & 2.73 \\
 
 & Test & 5.88 & 0.24 & 5.96 & 0.46 & 10.12 & 0.29 & 10.32 & 2.37 & 10.94 & 2.73 & 11.66 & 1.75 & 13.06 & 3.02 \\
 
\bottomrule
\end{tabular}
}
\caption{Average performance and standard errors for all models from Section 5.2 on the test set of 1,603 domain-general dictionary sentences and the validation datasets of 500 sentences from the Bible and 500 from the WT. Number of sentences in each training data setup shown in brackets.}
\label{tab:full_results_2}
\end{table*}


\newgeometry{top=1cm}
\begin{table}
\centering
\tiny
\begin{tabular}{p{1.3cm}p{2cm}llllll}
\toprule
\multicolumn{2}{l}{\textbf{Combined Tokeniser, Shared Embeddings}} & \multicolumn{6}{c}{Vocabulary Size} \\
& & \multicolumn{2}{c}{10000} & \multicolumn{2}{c}{20000} & \multicolumn{2}{c}{30000} \\
Language Pair & Evaluation Set & Avg & StdE & Avg & StdE & Avg & StdE  \\
\midrule
\multirow[c]{4}{*}{Kir-Eng} & Val bible & 19.68 & 0.52 & 18.96 & 0.31 & 16.80 & 1.62 \\
 & Val WT & 9.08 & 0.37 & 9.34 & 0.30 & 7.42 & 0.94 \\
 & Val all & 14.38 & 0.42 & 14.15 & 0.29 & 12.11 & 1.27 \\
 & Test & 10.14 & 0.96 & 11.46 & 0.24 & 8.86 & 1.40 \\
\cline{1-8}
\multirow[c]{4}{*}{Kir-Por} & Val bible & 19.16 & 0.96 & 17.16 & 2.47 & 17.5 & 1.19 \\
 & Val WT & 8.20 & 0.13 & 7.02 & 1.08 & 6.72 & 0.70 \\
 & Val all & 13.68 & 0.41 & 12.09 & 1.77 & 12.11 & 0.94 \\
 & Test & 13.06 & 0.60 & 8.62 & 2.03 & 7.36 & 1.56 \\
\cline{1-8}
\multirow[c]{4}{*}{Eng-Kir} & Val bible & 22.34 & 0.65 & 23.20 & 0.40 & 22.32 & 0.45 \\
 & Val WT & 9.02 & 0.17 & 9.74 & 0.19 & 9.18 & 0.28 \\
 & Val all & 15.68 & 0.36 & 16.47 & 0.25 & 15.75 & 0.29 \\
 & Test & 11.14 & 1.17 & 11.96 & 0.32 & 11.44 & 0.33 \\
\cline{1-8}
\multirow[c]{4}{*}{Por-Kir} & Val bible & 19.24 & 4.04 & 20.62 & 0.54 & 21.76 & 0.07 \\
 & Val WT & 7.76 & 1.45 & 7.94 & 0.21 & 8.7 & 0.17 \\
 & Val all & 13.50 & 2.73 & 14.28 & 0.27 & 15.23 & 0.10 \\
 & Test & 13.06 & 3.02 & 11.5 & 1.40 & 12.44 & 0.54 \\
\bottomrule
\end{tabular}

\begin{tabular}{p{1.3cm}p{2cm}llllll}
\toprule
\multicolumn{2}{l}{\textbf{Combined Tokeniser, Separate Embeddings}} & \multicolumn{6}{c}{Vocabulary Size} \\
& & \multicolumn{2}{c}{10000} & \multicolumn{2}{c}{20000} & \multicolumn{2}{c}{30000} \\
Language Pair & Evaluation Set & Avg & StdE & Avg & StdE & Avg & StdE  \\
\midrule
\multirow[c]{4}{*}{Kir-Eng} & Val bible & 17.92 & 1.37 & 13.64 & 2.23 & 15.28 & 3.47 \\
 & Val WT & 7.18 & 0.87 & 5.22 & 0.91 & 5.86 & 1.27 \\
 & Val all & 12.55 & 1.11 & 9.43 & 1.55 & 10.57 & 2.36 \\
 & Test & 7.52 & 1.42 & 5.64 & 1.35 & 5.78 & 1.53 \\
\cline{1-8}
\multirow[c]{4}{*}{Kir-Por} & Val bible & 12.12 & 1.86 & 15.88 & 2.36 & 14.40 & 1.88 \\
 & Val WT & 4.06 & 0.43 & 5.50 & 0.78 & 4.94 & 0.58 \\
 & Val all & 8.09 & 1.14 & 10.69 & 1.56 & 9.67 & 1.21 \\
 & Test & 3.46 & 0.83 & 5.56 & 1.29 & 4.34 & 1.05 \\
\cline{1-8}
\multirow[c]{4}{*}{Eng-Kir} & Val bible & 18.38 & 2.92 & 19.68 & 1.10 & 16.58 & 2.62 \\
 & Val WT & 6.48 & 1.01 & 7.34 & 10.56 & 5.66 & 1.10 \\
 & Val all & 12.43 & 1.96 & 13.51 & 0.81 & 11.12 & 1.85 \\
 & Test & 6.38 & 1.25 & 7.08 & 1.04 & 5.80 & 1.56 \\
\cline{1-8}
\multirow[c]{4}{*}{Por-Kir} & Val bible & 13.62 & 2.17 & 12.96 & 3.13 & 13.9 & 2.12 \\
 & Val WT & 4.66 & 0.67 & 4.52 & 0.95 & 4.64 & 0.60 \\
 & Val all & 9.14 & 1.41 & 8.74 & 2.04 & 9.27 & 1.36 \\
 & Test & 4.58 & 1.34 & 3.88 & 1.42 & 4.24 & 0.90 \\

\bottomrule
\end{tabular}

\begin{tabular}{p{1.3cm}p{2cm}llllll}
\toprule
\multicolumn{2}{l}{\textbf{Separate Tokeniser, Shared Embeddings}} & \multicolumn{6}{c}{Vocabulary Size} \\
& & \multicolumn{2}{c}{10000} & \multicolumn{2}{c}{20000} & \multicolumn{2}{c}{30000} \\
Language Pair & Evaluation Set & Avg & StdE & Avg & StdE & Avg & StdE  \\
\midrule
\multirow[c]{4}{*}{Kir-Eng} & Val bible & 15.28 & 0.42 & 17.22 & 0.37 & 14.32 & 2.98 \\
 & Val WT & 2.24 & 0.12 & 2.42 & 0.06 & 2.16 & 0.44 \\
 & Val all & 8.76 & 0.24 & 9.82 & 0.18 & 8.24 & 1.71 \\
 & Test & 10.82 & 0.15 & 10.36 & 0.50 & 9.1 & 2.15 \\
\cline{1-8}
\multirow[c]{4}{*}{Kir-Por} & Val bible & 15.88 & 0.69 & 17.24 & 0.44 & 17.76 & 0.20 \\
 & Val WT & 2.22 & 0.04 & 2.9 & 0.06 & 2.66 & 0.19 \\
 & Val all & 9.05 & 0.34 & 10.07 & 0.20 & 10.21 & 0.15 \\
 & Test & 13.00 & 0.24 & 12.42 & 0.33 & 11.92 & 0.29 \\
\cline{1-8}
\multirow[c]{4}{*}{Eng-Kir} & Val bible & 20.24 & 0.51 & 20.08 & 0.87 & 20.14 & 0.37 \\
 & Val WT & 2.78 & 0.15 & 2.86 & 0.10 & 2.76 & 0.24 \\
 & Val all & 11.51 & 0.31 & 11.47 & 0.47 & 11.45 & 0.26 \\
 & Test & 11.84 & 0.82 & 12.78 & 0.24 & 12.58 & 0.47 \\
\cline{1-8}
\multirow[c]{4}{*}{Por-Kir} & Val bible & 17.76 & 3.04 & 18.00 & 3.95 & 16.98 & 3.49 \\
 & Val WT & 2.22 & 0.44 & 2.42 & 1.48 & 2.60 & 0.54 \\
 & Val all & 9.99 & 1.74 & 10.21 & 2.22 & 9.79 & 2.01 \\
 & Test & 12.52 & 2.83 & 14.08 & 3.43 & 12.5 & 3.00 \\
\bottomrule
\end{tabular}

\begin{tabular}{p{1.3cm}p{2cm}llllll}
\toprule
\multicolumn{2}{l}{\textbf{Separate Tokeniser, Separate Embeddings}} & \multicolumn{6}{c}{Vocabulary Size} \\
& & \multicolumn{2}{c}{10000} & \multicolumn{2}{c}{20000} & \multicolumn{2}{c}{30000} \\
Language Pair & Evaluation Set & Avg & StdE & Avg & StdE & Avg & StdE  \\
\midrule
\multirow[c]{4}{*}{Kir-Eng} & Val bible & 19.78 & 0.62 & 15.64 & 1.93 & 19.62 & 0.86 \\
 & Val WT & 7.50 & 0.18 & 5.72 & 0.77 & 7.42 & 0.28 \\
 & Val all & 13.64 & 0.34 & 10.68 & 1.34 & 13.52 & 0.51 \\
 & Test & 8.22 & 0.43 & 5.94 & 1.16 & 8.66 & 0.52 \\
\cline{1-8}
\multirow[c]{4}{*}{Kir-Por} & Val bible & 17.32 & 1.17 & 13.22 & 2.39 & 16.94 & 0.61 \\
 & Val WT & 5.16 & 0.48 & 4.52 & 0.72 & 5.2 & 0.39 \\
 & Val all & 11.24 & 0.82 & 8.87 & 1.55 & 11.07 & 0.44 \\
 & Test & 6.30 & 0.91 & 5.00 & 1.48 & 5.72 & 0.81 \\
\cline{1-8}
\multirow[c]{4}{*}{Eng-Kir} & Val bible & 17.36 & 2.14 & 15.30 & 3.71 & 12.82 & 4.24 \\
 & Val WT & 6.64 & 0.80 & 6.08 & 1.27 & 5.00 & 1.48 \\
 & Val all & 12.00 & 1.46 & 10.69 & 2.48 & 8.91 & 2.86 \\
 & Test & 5.78 & 1.19 & 4.72 & 1.45 & 4.54 & 1.87 \\
\cline{1-8}
\multirow[c]{4}{*}{Por-Kir} & Val bible & 17.80 & 2.80 & 14.94 & 4.31 & 15.06 & 3.77 \\
 & Val WT & 5.94 & 0.92 & 5.5 & 1.28 & 4.94 & 1.28 \\
 & Val all & 11.87 & 1.86 & 10.22 & 2.79 & 10.00 & 2.52 \\
 & Test & 6.90 & 1.64 & 6.64 & 2.42 & 6.08 & 2.15 \\
\bottomrule
\end{tabular}

\caption{Average performance and standard errors for all models from Section 5.3 on the test set of 1,603 domain-general dictionary sentences and the validation datasets of 500 sentences from the Bible and 500 from the WT.}
\label{tab:full_results_3}
\end{table}

\begin{table}
\centering
\tiny
\begin{tabular}{p{1.3cm}p{2cm}llllll}
\toprule
\multicolumn{2}{l}{\textbf{Combined Tokeniser, Shared Embeddings}} & \multicolumn{6}{c}{Vocabulary Size} \\
& & \multicolumn{2}{c}{10000} & \multicolumn{2}{c}{20000} & \multicolumn{2}{c}{30000} \\
Language Pair & Evaluation Set & Avg & StdE & Avg & StdE & Avg & StdE  \\
\midrule
\multirow[c]{4}{*}{Kir-Eng} & Val bible & 19.68 & 0.52 & 18.96 & 0.31 & 19.40 & 0.55 \\
 & Val WT & 9.08 & 0.37 & 9.34 & 0.30 & 8.87 & 0.39 \\
 & Val all & 14.38 & 0.42 & 14.15 & 0.29 & 14.13 & 0.47 \\
 & Test & 10.14 & 0.96 & 11.46 & 0.24 & 10.73 & 0.09 \\
\cline{1-8}
\multirow[c]{4}{*}{Kir-Por} & Val bible & 19.16 & 0.96 & 19.63 & 0.26 & 19.17 & 0.61 \\
 & Val WT & 8.20 & 0.13 & 8.10 & 0.12 & 7.80 & 0.15 \\
 & Val all & 13.68 & 0.41 & 13.86 & 0.09 & 13.48 & 0.38 \\
 & Test & 13.06 & 0.60 & 10.60 & 0.57 & 9.67 & 0.28 \\
\cline{1-8}
\multirow[c]{4}{*}{Eng-Kir} & Val bible & 22.34 & 0.65 & 23.20 & 0.40 & 22.32 & 0.45 \\
 & Val WT & 9.02 & 0.17 & 9.74 & 0.19 & 9.18 & 0.28 \\
 & Val all & 15.68 & 0.36 & 16.47 & 0.25 & 15.75 & 0.29 \\
 & Test & 11.14 & 1.17 & 11.96 & 0.32 & 11.44 & 0.33 \\
\cline{1-8}
\multirow[c]{4}{*}{Por-Kir} & Val bible & 23.23 & 0.83 & 20.62 & 0.54 & 21.76 & 0.07 \\
 & Val WT & 9.20 & 0.18 & 7.94 & 0.21 & 8.70 & 0.17 \\
 & Val all & 16.21 & 0.41 & 14.28 & 0.27 & 15.23 & 0.10 \\
 & Test & 15.98 & 1.01 & 11.50 & 1.40 & 12.44 & 0.54 \\
\bottomrule
\end{tabular}

\begin{tabular}{p{1.3cm}p{2cm}llllll}
\toprule
\multicolumn{2}{l}{\textbf{Combined Tokeniser, Separate Embeddings}} & \multicolumn{6}{c}{Vocabulary Size} \\
& & \multicolumn{2}{c}{10000} & \multicolumn{2}{c}{20000} & \multicolumn{2}{c}{30000} \\
Language Pair & Evaluation Set & Avg & StdE & Avg & StdE & Avg & StdE  \\
\midrule
\multirow[c]{4}{*}{Kir-Eng} & Val bible & 19.20 & 0.63 & 17.85 & 1.05 & 18.68 & 0.90 \\
 & Val WT & 7.93 & 0.57 & 7.30 & 0.30 & 7.08 & 0.49 \\
 & Val all & 13.56 & 0.60 & 12.58 & 0.68 & 12.88 & 0.65 \\
 & Test & 8.65 & 1.11 & 8.75 & 0.65 & 7.20 & 0.74 \\
 
\cline{1-8}
\multirow[c]{4}{*}{Kir-Por} & Val bible & 16.20 & 1.40 & 18.13 & 0.93 & 16.23 & 0.58 \\
 & Val WT & 5.05 & 0.25 & 6.23 & 0.38 & 5.48 & 0.29 \\
 & Val all & 10.63 & 0.82 & 12.18 & 0.63 & 10.85 & 0.35 \\
 & Test & 5.45 & 0.07 & 6.70 & 0.78 & 5.20 & 0.77 \\
\cline{1-8}
\multirow[c]{4}{*}{Eng-Kir} & Val bible & 21.25 & 0.66 & 21.47 & 0.19 & 22.05 & 0.15 \\
 & Val WT & 7.45 & 0.38 & 8.17 & 0.23 & 8.20 & 0.00 \\
 & Val all & 14.35 & 0.51 & 14.82 & 0.17 & 15.13 & 0.07 \\
 & Test & 7.55 & 0.58 & 8.27 & 1.22 & 9.20 & 0.90 \\
\cline{1-8}
\multirow[c]{4}{*}{Por-Kir} & Val bible & 18.50 & 2.00 & 19.50 & 0.50 & 18.60 & 1.30 \\
 & Val WT & 6.00 & 1.10 & 6.50 & 0.60 & 6.05 & 0.25 \\
 & Val all & 12.25 & 1.55 & 13.00 & 0.55 & 12.33 & 0.77 \\
 & Test & 7.20 & 2.40 & 7.05 & 1.05 & 6.10 & 1.50 \\

\bottomrule
\end{tabular}

\begin{tabular}{p{1.3cm}p{2cm}llllll}
\toprule
\multicolumn{2}{l}{\textbf{Separate Tokeniser, Shared Embeddings}} & \multicolumn{6}{c}{Vocabulary Size} \\
& & \multicolumn{2}{c}{10000} & \multicolumn{2}{c}{20000} & \multicolumn{2}{c}{30000} \\
Language Pair & Evaluation Set & Avg & StdE & Avg & StdE & Avg & StdE  \\
\midrule
\multirow[c]{4}{*}{Kir-Eng} & Val bible & 15.28 & 0.42 & 17.22 & 0.37 & 17.28 & 0.51 \\
 & Val WT & 2.24 & 0.12 & 2.42 & 0.06 & 2.58 & 0.21 \\
 & Val all & 8.76 & 0.24 & 9.82 & 0.18 & 9.93 & 0.34 \\
 & Test & 10.82 & 0.15 & 10.36 & 0.50 & 11.23 & 0.45 \\
\cline{1-8}
\multirow[c]{4}{*}{Kir-Por} & Val bible & 15.88 & 0.69 & 17.24 & 0.44 & 17.76 & 0.20 \\
 & Val WT & 2.22 & 0.04 & 2.9 & 0.06 & 2.66 & 0.19 \\
 & Val all & 9.05 & 0.34 & 10.07 & 0.20 & 10.21 & 0.15 \\
 & Test & 13.00 & 0.24 & 12.42 & 0.33 & 11.92 & 0.29 \\
\cline{1-8}
\multirow[c]{4}{*}{Eng-Kir} & Val bible & 20.24 & 0.51 & 20.08 & 0.87 & 20.14 & 0.37 \\
 & Val WT & 2.78 & 0.15 & 2.86 & 0.10 & 2.76 & 0.24 \\
 & Val all & 11.51 & 0.31 & 11.47 & 0.47 & 11.45 & 0.26 \\
 & Test & 11.84 & 0.82 & 12.78 & 0.24 & 12.58 & 0.47 \\
\cline{1-8}
\multirow[c]{4}{*}{Por-Kir} & Val bible & 20.78 & 0.51 & 21.95 & 0.25 & 20.45 & 0.42 \\
 & Val WT & 2.65 & 0.10 & 2.90 & 0.04 & 3.13 & 0.14 \\
 & Val all & 11.71 & 0.30 & 12.43 & 0.10 & 11.79 & 0.23 \\
 & Test & 15.33 & 0.49 & 17.50 & 0.27 & 15.40 & 1.00 \\
\bottomrule
\end{tabular}

\begin{tabular}{p{1.3cm}p{2cm}llllll}
\toprule
\multicolumn{2}{l}{\textbf{Separate Tokeniser, Separate Embeddings}} & \multicolumn{6}{c}{Vocabulary Size} \\
& & \multicolumn{2}{c}{10000} & \multicolumn{2}{c}{20000} & \multicolumn{2}{c}{30000} \\
Language Pair & Evaluation Set & Avg & StdE & Avg & StdE & Avg & StdE  \\
\midrule
\multirow[c]{4}{*}{Kir-Eng} & Val bible & 19.78 & 0.62 & 19.45 & 1.85 & 19.62 & 0.86 \\
 & Val WT & 7.50 & 0.18 & 7.35 & 0.95 & 7.42 & 0.28 \\
 & Val all & 13.64 & 0.34 & 13.40 & 1.40 & 13.52 & 0.51 \\
 & Test & 8.22 & 0.43 & 8.55 & 0.75 & 8.66 & 0.52 \\
\cline{1-8}
\multirow[c]{4}{*}{Kir-Por} & Val bible & 18.35 & 0.72 & 18.00 & 1.00 & 16.94 & 0.61 \\
 & Val WT & 5.58 & 0.31 & 6.00 & 0.50 & 5.20 & 0.39 \\
 & Val all & 11.96 & 0.50 & 12.00 & 0.75 & 11.07 & 0.44 \\
 & Test & 6.95 & 0.83 & 7.95 & 2.15 & 5.72 & 0.81 \\
\cline{1-8}
\multirow[c]{4}{*}{Eng-Kir} & Val bible & 20.83 & 0.33 & 22.10 & 0.60 & 21.50 & 1.10 \\
 & Val WT & 7.93 & 0.20 & 8.60 & 0.60 & 8.15 & 0.55 \\
 & Val all & 14.38 & 0.07 & 15.35 & 0.60 & 14.83 & 0.83 \\
 & Test & 7.67 & 0.47 & 7.85 & 0.15 & 8.70 & 0.20 \\
\cline{1-8}
\multirow[c]{4}{*}{Por-Kir} & Val bible & 20.53 & 0.85 & 21.93 & 0.70 & 19.43 & 0.73 \\
 & Val WT & 6.80 & 0.44 & 7.57 & 0.33 & 6.57 & 0.27 \\
 & Val all & 13.66 & 0.64 & 14.75 & 0.48 & 13.00 & 0.49 \\
 & Test & 8.30 & 1.10 & 10.50 & 0.96 & 8.57 & 2.42 \\
\bottomrule
\end{tabular}

\caption{Average performance and standard errors for selected models from Section 5.3 on the test set of 1,603 domain-general dictionary sentences and the validation datasets of 500 sentences from the Bible and 500 from the WT. Models with BLEU score on the validation set of more than 3.0 BLEU lower than the highest-scoring model on the validation set in that group have been removed.}
\label{tab:full_results_4}
\end{table}
\restoregeometry

\section{Human validation}
\label{app:hval}

\begin{figure*}[h]
\centering
\vspace{-180mm}
\includegraphics[width=1\linewidth]{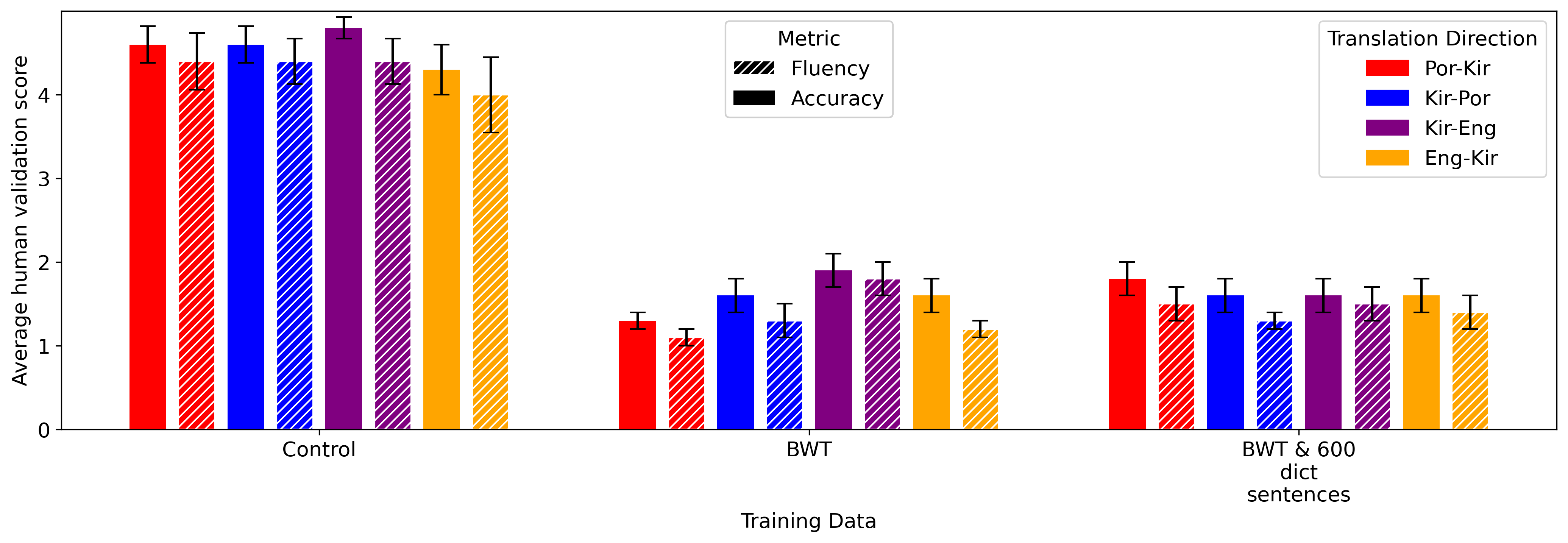}
\caption{Average scores of human judgements for accuracy (solid) and fluency (hatched) of translated sentences from the reference sets (control) and from models
trained on Bible and WT data (BWT) and Bible, WT and 600 dictionary sentences.
Standard errors across model sets are shown with error bars.}
\label{fig:human_val_2}
\end{figure*}

We recruited one native speaker of creole who is fluent in Portuguese and English to evaluate sample sentences from models trained on the Bible and WT data alone, and models trained on the Bible and WT data with 600 sentences from the dictionary. We selected 25 sentences from the test set, and then used both models in each of the four language directions (Portuguese-Kiriol, Kiriol-Portuguese, English-Kiriol, Kiriol-English) to translate the selected sentences for the participant to evaluate. This provided 50 sentences for each language direction, 25 from the model with dictionary data included and 25 from the model without. For each language direction, we also included 10 translations from the reference set, to serve as a control measure and check the instructions were being followed appropriately. This resulted in 60 sentences per language direction for the participant to evaluate, so 240 sentences overall. We expected the task to take between 2 and 4 hours, allocating £76 to give an hourly fee of £19-£38 per hour depending on the participant's speed. The task was carried out via a Qualtrics survey. The instructions are included below, and \Cref{fig:human_val_2} shows the average judgements for each language direction over all three conditions. \\

\textbf{Instructions}

\setlength{\parindent}{0pt}
In this questionnaire, you'll be presented with 240 pairs of sentences, with the first sentence as the original sentence and the second sentence as the translated sentence. There are translations from:

\begin{enumerate}
\setlength{\itemsep}{-2pt}
    \item Kiriol into English
    \item Kiriol into Portuguese
    \item English into Kiriol
    \item Portuguese into Kiriol
\end{enumerate}

The sentences and their translations will be presented like this: \\

Source: A chuva trouxe alegria na aldeia.\\
Translation: I manda cuba na tabanka.\\

For each question, first read the source and the translation sentence carefully. You will then be asked to rate the translation for its adequacy (how much of the meaning of the source sentence it captures) and its fluency (how natural a sentence it is in the target language). You can give a rating of 1-5 for each metric by sliding the marker. You should think of the scores as follows:\\

\textbf{Adequacy:}
\begin{enumerate}
\setlength{\itemsep}{-2pt}
    \item [5.] All Meaning
    \item [4.] Most Meaning
    \item [3.] Much Meaning
    \item [2.] Little Meaning
    \item [1.] None
\end{enumerate}

\textbf{Fluency:}
\begin{enumerate}
\setlength{\itemsep}{-2pt}
    \item [5.] Flawless
    \item [4.] Good
    \item [3.] Non-native
    \item [2.] Disfluent
    \item [1.] Incomprehensible
\end{enumerate}

You might want to keep a copy of these guidelines to hand as you answer the questions. \\

A sentence might be adequate in terms of meaning, but not very fluent if it doesn't sound natural or have the correct grammar. Conversely, a sentence might be very fluent but have very little to do with the source sentence, rendering it inaccurate. \\

Please provide a score for both adequacy and fluency for each sentence. Some of the source sentences are the same, but the translations will always be slightly different.\\

We will also provide a comment box at the bottom of each page in case you wish to comment on specific examples or word choices that you think are interesting, or for example if one of the translations is rude or culturally insensitive (please list the relevant question number next to your comment). There is no obligation to provide comments unless you wish to. \\

Your answers will be saved by Qualtrics and so you can take breaks when you need to and come back to finish at a later time. We anticipate that you should be able to complete all 240 sentences within 4 hours (roughly 1 minute per sentence).
\\ 
If you have any questions about the instructions or about specific examples, please do not hesitate to contact jacqueline.rowe@ed.ac.uk for further guidance.

\end{document}